\documentclass[acmsmall]{acmart}
\usepackage{tikz}
\usepackage{tikz-cd}
\usetikzlibrary{automata,arrows,shapes,fit, positioning, calc, decorations.pathreplacing}
\usepackage{ifthen} 
\usepackage{algorithm}
\usepackage{algpseudocode}
\usepackage{float}
\usepackage{microtype}
\usepackage{wrapfig}
\usepackage{placeins}
\usepackage{adjustbox}
\usepackage{booktabs}
\usepackage{pgfplots}
\pgfplotsset{compat=newest}
\usepackage{siunitx}
\usepackage{multirow}
\definecolor{lightgreen}{RGB}{220, 250, 220}
\definecolor{lightred}{RGB}{255, 230, 230}
\usepackage{graphicx}
\usepackage{enumitem}
\usepackage{subcaption}
\usepackage{upgreek}
\usepackage{amsmath} 
\usepackage{amsthm}
\newtheorem{definition}{Definition}
\newtheorem{remark}{Remark}

\AtBeginDocument{
  }

\setcopyright{cc}
\setcctype{by}
\acmDOI{10.1145/3808203}
\acmYear{2026}
\acmJournal{PACMSE}
\acmVolume{3}
\acmNumber{FSE}
\acmArticle{FSE196}
\acmMonth{7}
\acmSubmissionID{fse26mainb-p2516-p}

\newcommand{\PP}{\mathtt{P}}
\newcommand{\RR}{\mathtt{R}}

\newcommand{\acronym}{SHARP}

\begin{document}

\title{Accelerating Policy Synthesis in Large-Scale MDPs via Hierarchical Adaptive Refinement}

\author{Alexandros Evangelidis}
\authornote{Alexandros Evangelidis is the corresponding author.}
\orcid{0000-0003-4032-3042}
\email{alexandros.evangelidis@york.ac.uk}
\affiliation{
  \institution{University of York}
  \city{York}
  \country{United Kingdom}
}

\author{Gricel Vázquez}
\orcid{0000-0003-4886-5567}
\email{gricel.vazquez@york.ac.uk}
\affiliation{
  \institution{University of York}
  \city{York}
  \country{United Kingdom}
}

  \author{Simos Gerasimou}
  \orcid{0000-0002-2706-5272}
  \email{simos.gerasimou@york.ac.uk}
  \affiliation{
    \institution{Cyprus University of Technology}
    \city{Limassol}
    \country{Cyprus}
  }
  \affiliation{
    \institution{University of York}
    \city{York}
    \country{United Kingdom}
  }

\renewcommand{\shortauthors}{Evangelidis, Vázquez, and Gerasimou}

\begin{abstract} 
Software-intensive systems, such as software product lines and robotics, utilise Markov decision processes (MDPs) to capture uncertainty and analyse sequential decision-making problems. 
Despite the usefulness of conventional policy synthesis methods, they fail to scale to large state spaces. 
Our approach addresses this issue and accelerates policy synthesis in large MDPs by dynamically refining the MDP and iteratively selecting the most fragile MDP regions for refinement. 
This iterative procedure offers a balance between accuracy and efficiency, as refinement occurs only when necessary.
We formally show that the composed policy is near-optimal under standard assumptions, with error bounded by the local solver tolerance and boundary mismatch.
Across diverse case studies and MDPs up to 1M states, we demonstrate that our approach achieves up to 2$\times$ speedup over PRISM, offering a competitive solution for real-world policy synthesis in large MDPs.
\end{abstract}

\ccsdesc[500]{Theory of computation~Markov decision processes}
\ccsdesc[300]{Computing methodologies~Planning under uncertainty}
 \keywords{Probabilistic Model Checking, Policy Synthesis, Hierarchical Refinement}

\maketitle

\section{Introduction}
\label{sec:into}
Modern software-intensive systems range from cloud microservices with frequent deployments~\cite{pujol2023edge,jawaddi2022review} to self-adaptive systems and autonomous robots navigating dynamic warehouses~\cite{wurman2007robots,zhao2019probabilistic,fisher2021overview}. These systems must reason rigorously about uncertainty~\cite{hezavehi2021uncertainty,weyns2023towards}. Markov decision processes (MDPs) capture both nondeterminism and stochasticity, making them an expressive mathematical framework for sequential decision-making under uncertainty~\cite{puterman1994mdps}.
In software engineering (SE), MDPs support the continuous verification of self-adaptive systems~\cite{su2016iterative,calinescu2012self,calinescu2018using}, dynamic QoS management and optimisation of service-based applications~\cite{calinescu2011qos}, variability-aware model checking of software product lines~\cite{profeat2016}, and
robotic software stacks for task and motion planning~\cite{survey2023robotics}.

 However, despite the widespread adoption of MDPs to support decision-making in these complex, software-intensive systems, their practical application is often hindered by the so-called state-space explosion problem~\cite{clarke2012explosion}.
 This explosion in the number of states and transitions renders standard policy synthesis algorithms, which determine optimal actions, computationally prohibitive due to excessive time and memory requirements~\cite{su2016iterative,baier2008principles}.
In probabilistic model checking, policy (strategy) synthesis is a verification task that not only checks a quantitative specification but also returns an executable controller (policy) that optimises it \cite{kwiatkowska2022pmcautonomy}.

For engineering teams that rely on MDP-based analysis to ensure safety requirements (e.g., ``the robot never becomes stuck'') or the achievement of performance targets (e.g., ``the task must be completed within $T$ time steps''), these scalability and efficiency limitations can significantly undermine the adoption of verification techniques for real-world software systems~\cite{kwiatkowska2022pmcautonomy}.
Large models can stall verification pipelines and delay releases. For example, even adjusting a single parameter (such as shelf layout) might force re-verification of the entire MDP, escalating development costs and turnaround time~\cite{junges2022abstraction,calinescu2018efficient}.

Despite significant research, the scalability of MDP solution methods remains a fundamental challenge~\cite{baier201910}. Recent benchmarks~\cite{budde2021qcomp} report that PRISM~\cite{prism} and Storm~\cite{storm} still time out on large instances, whether using symbolic~\cite{alfaro2000symbolic}, explicit~\cite{hartmanns2015explicit}, or sampling-based~\cite{henriques2012statistical} backends. 

To address this challenge, we introduce \textbf{SHARP}, an approach for \textbf{S}calable \textbf{H}ierarchical \textbf{A}daptive \textbf{R}efinement for \textbf{P}olicy synthesis.
The core of our approach is based upon a ``refine-where-needed'' principle, which focuses computational effort only on the most critical regions of the state space. 
\acronym\ begins with a coarse partitioning of the MDP and iteratively improves it by subdividing and solving state space partitions, yielding a refined and more detailed analysis.
Specifically, \acronym\ hierarchically partitions the state space, solves sub-MDPs with boundary values, and refines blocks whose spread exceeds a depth-adapted threshold, since low-spread blocks are already well-resolved.

  Partition-refinement has a long history in probabilistic verification:
  CEGAR~\cite{clarke2000cegar,hermanns2008probabilistic,kattenbelt2010game} refines abstractions to eliminate spurious counterexamples,
  bisimulation-based methods~\cite{ferns2004metrics} aggregate states with similar dynamics into abstract states,
  and parametric abstraction-refinement~\cite{junges2022abstraction} analyses hierarchical MDPs.
  These techniques are primarily \emph{verification-driven}: they are designed to establish quantitative bounds and may employ state aggregation as long as the bound is preserved.
  Policy synthesis is more demanding, because it must commit to an action choice at each concrete state; aggregating states that require different actions can of course degrade the resulting controller~\cite{baier2008principles} (though not necessarily a fatal degradation).
  In a partitioned solve, boundary values between partitions evolve across iterations.
  Consequently, the locally optimal action can change, potentially degrading the composed global policy unless boundary drift is controlled (see Section~\ref{subsec:sharp-theory}).
  \acronym\ addresses these challenges because it operates on the concrete state space (without merging states) and refines blocks based on value spread and not on counterexample elimination.
  We also show that, provided the boundary drift is controlled, the composed global policy remains near-optimal.

  Our contributions can be summarised as follows. We propose \acronym, a scalable
  policy-synthesis method based upon hierarchical adaptive refinement, with block-based
  selection, boundary-aware sub-MDPs, and bottom-up propagation. Furthermore, we provide
  formal guarantees on convergence, termination, and error under standard assumptions.
  Moreover, we implement the approach as part of the PRISM model checker, with configurable
  refinement strategies. Finally, we evaluate \acronym\ on a diverse set of MDPs, including models with up to 1M states, and show speedups of up to 2$\times$ over PRISM's explicit-state engine on large structured instances.

The rest of the paper is organised as follows. Sections~\ref{sec:prelims}--\ref{sec:motiv-example} provide background and a motivating example. Section~\ref{sec:approach} presents the \acronym\ approach with its theoretical guarantees, followed by the implementation (Section~\ref{sec:implementation}) and evaluation (Section~\ref{sec:evaluation}). Sections~\ref{sec:relatedwork}--\ref{sec:conclusion} discuss related work and conclude the paper.

\section{Preliminaries}
\label{sec:prelims}

\subsection{Probabilistic Modelling}
\textbf{Markov decision process (MDP)}. 
An MDP is a tuple
\(\mathcal{M} \;=\; (S, \bar{s}, A, \delta, c, AP, L)\), where $S$ is a finite set of states; $\bar{s}\in S$ is an initial state; $A$ is a finite set of actions; $\delta: (S \times A)\to \mathit{Dist}(S)$ is a probabilistic transition function mapping each $(s,a)$ pair to a probability distribution over $S$;
\(c : S \times A \to \mathbb R_{\ge 0}\) is a (per-action) cost function;
$AP$ is a set of atomic propositions; and $L: S \to 2^{AP}$ is a labelling function.
Intuitively, states are configurations, actions are controllable choices, and $\delta$ gives probabilistic transitions, while costs encode penalties.

MDPs support nondeterministic actions as each state $s\in S$ can have multiple enabled actions for which $\delta(s,a)$ is defined, given by $A(s)$. 
An MDP \textit{policy} is a possible resolution of nondeterminism. 
Resolving this nondeterminism allows us to probabilistically reason about MDPs and synthesise \textit{memoryless deterministic policies}, i.e., policies where the action chosen in a state depends solely on the current state. 
Formally, a (deterministic memoryless) policy of an MDP is a function $\sigma :S \rightarrow A$ mapping each state $s\in S$ to an action in $A(s)$.

\textbf{Stochastic shortest path (SSP) MDP}.
Let $\mathcal{M}$ be the MDP defined above and $G\subseteq S$ be a set of goal states.
We say $\mathcal{M}$ is an SSP instance \cite{bertsekas1991ssp} if:
(i) goals are absorbing and carry zero cost, i.e., for all $g\in G$ and $a\in A(g)$ we have $\delta(g,a)(g)=1$ and $c(g,a)=0$;
(ii) there exists a \emph{proper} policy that reaches $G$ with probability $1$ from the states of interest; and
(iii) every improper policy (one that reaches $G$ with probability $<1$ from some state) has infinite expected total cost from at least one state.
We use non-negative per-action costs $c(s,a)\!\ge\!0$, accumulated \emph{until the first visit to~$G$}.
From an SSP point of view, the task is to reach the goal while accounting for the cost incurred before the first visit to $G$.
We use PCTL for the probabilistic reachability objectives studied in the paper, and reward operators for the cost objectives, following the standard definitions~\cite{hansson1994pctl,bianco1995model}.

\subsection{Probabilistic Model Checking}
\label{subsec:pmc}
Probabilistic model checking (PMC)~\cite{kwiatkowska2007stochastic} computes quantitative reachability/cost objectives on stochastic models (e.g., MDPs), using tools such as PRISM~\cite{prism} and Storm~\cite{storm}.

\textbf{Quantitative reachability.}
Let $\mathcal{M}=(S,\bar{s},A,\delta,c,AP,L)$ be the MDP defined above and
$G\subseteq S$ a set of target states.
We denote by $\mathsf{F}\,G$ the event that $G$ is eventually reached.
For any state \(s\in S\) and policy \(\sigma\), let \(\Pr^{\sigma}_{s}(\cdot)\) and \(\mathbb{E}^{\sigma}_{s}[\cdot]\) denote probability and expectation over infinite paths of the Markov chain induced by \(\sigma\) from \(s\).
The maximum reachability probability is \(\PP_{\max}(s,\mathsf{F}\,G)=\sup_{\sigma}\Pr^\sigma_s(\mathsf{F}\,G)\).
The minimum expected cumulative cost to reach $G$ is
\(\RR_{\min}(s,\mathsf{F}G) = \inf_{\sigma} \mathbb{E}^{\sigma}_{s}\bigl[\sum_{k<k_G} c(\omega_k, a_k)\bigr]\),
where $a_k=\sigma(\omega_k)$ and
\(k_G=\min\{j \mid \omega_j\in G\}\) is the first index at which a state in $G$ is reached.

In other words, $\PP_{\max}$ asks for the maximum probability of reaching the goal over all policies, while $\RR_{\min}$ asks for the minimum expected cumulative cost to do so.
In the rest of the paper, we focus on $\PP_{\max}$ and $\RR_{\min}$, which match our use cases and experiments.

\begin{wrapfigure}[13]{R}{0.30\columnwidth}
\centering
\begin{tikzpicture}[scale=0.55, font=\footnotesize]
  \useasboundingbox (-0.1,-0.1) rectangle (6.1,6.3);

  \fill[gray!10] (0,0) rectangle (6,6);
  \draw[step=1.0, gray!40, thin] (0,0) grid (6,6);
  \draw[gray!70, line width=1.2pt] (0,0) rectangle (6,6);

  \fill[black!60] (1,2) rectangle (2,4);
  \draw[black, line width=0.8pt] (1,2) rectangle (2,4);
  \node[font=\scriptsize, fill=white, rounded corners=2pt, inner sep=1pt]
    at (1.5,4.2) {\textbf{W1}};

  \fill[black!60] (4,1) rectangle (5,3);
  \draw[black, line width=0.8pt] (4,1) rectangle (5,3);
  \node[font=\scriptsize, fill=white, rounded corners=2pt, inner sep=1pt]
    at (4.5,3.2) {\textbf{W2}};

  \fill[black!60] (2,5) rectangle (4,6);
  \draw[black, line width=0.8pt] (2,5) rectangle (4,6);
  \node[font=\scriptsize, fill=white, rounded corners=2pt, inner sep=1pt]
    at (3,4.8) {\textbf{W3}};

  \fill[green!40, opacity=0.5] (0,0) rectangle (1,1);
  \draw[green!70!black, line width=0.8pt] (0,0) rectangle (1,1);
  \node[font=\scriptsize, green!40!black] at (0.5,1.25) {Start};

  \begin{scope}[shift={(0.5,0.5)}]
    \fill[gray!80] (0,0) circle (0.225);
    \fill[cyan!80] (-0.075,0.075) circle (0.05);
    \fill[cyan!80] (0.075,0.075) circle (0.05);
    \draw[black, line width=1pt] (0,0.225) -- (0,0.33);
    \fill[red] (0,0.33) circle (0.04);
  \end{scope}

  \fill[blue!40, opacity=0.5] (5,5) rectangle (6,6);
  \draw[blue!70!black, line width=0.8pt] (5,5) rectangle (6,6);
  \node[font=\scriptsize, blue!40!black] at (5.5,4.8) {Goal};

  \fill[red!20, opacity=0.2] (0.8,1.8) rectangle (2.2,4.2);
  \fill[red!20, opacity=0.2] (3.8,0.8) rectangle (5.2,3.2);
  \fill[red!20, opacity=0.2] (1.8,4.8) rectangle (4.2,6.2);
  \node[font=\tiny, red!60!black] at (1.5,4.45) {Uncertain};
  \node[font=\tiny, red!60!black] at (4.5,3.5) {Uncertain};
  \node[font=\tiny, red!60!black] at (3.8,4.55) {Uncertain};

  \foreach \x in {0,...,5} {
    \node[font=\tiny, gray!70] at (\x+0.2,0.2) {\x};
  }
  \foreach \y in {0,...,5} {
    \node[font=\tiny, gray!70] at (0.2,\y+0.2) {\y};
  }
\end{tikzpicture}
\caption{A warehouse layout with shelves W1--W3.}
\label{fig:warehouse-fig}
\Description{A compact warehouse grid with a start area, a goal area, and three shelves labelled W1, W2, and W3. Regions near the shelves are marked uncertain.}
\end{wrapfigure}

\textbf{Value Iteration (VI).}
We use VI \cite{puterman1994mdps,chatterjee2008iteration} as the underlying
dynamic-programming procedure to compute both reachability probabilities and
expected costs. The method repeatedly applies the Bellman operator to update
the current value estimates. For example, after \(k\) iterations,
\(V^{(k)}(s)\) and \(W^{(k)}(s)\) denote the current values for state \(s\) in
the probability and cost cases, respectively. For each state \(s\), these
converge to \(\PP_{\max}(s,\mathsf{F}\,G)\) and
\(\RR_{\min}(s,\mathsf{F}\,G)\), respectively. Gauss-Seidel (GS) follows the same principle,
but it uses the newest available value estimates within an iteration, leading to faster convergence in
practice~\cite{baier2008principles}.

\section{Motivating Example: Warehouse Robotics}
\label{sec:motiv-example}
\noindent
Consider an autonomous robot operating in a warehouse with multiple shelves (W1--W3 in Figure~\ref{fig:warehouse-fig}), each possibly blocking certain aisles.
The robot should collect items from these shelves and deliver them to a packing station (Goal).
This environment is typically modelled as an MDP, where each state captures the robot's position and which aisles are blocked, while actions enable the robot to move up, down, left, or right.
Since the robot does not always move precisely as intended, each action leads to multiple possible next states with known probabilities (e.g., it might fail to move or slip to an adjacent cell).
This probabilistic behaviour, combined with large grids of shelves, yields MDPs with millions of states, rendering policy synthesis prohibitively expensive.

This challenging scenario frequently arises in large-scale, realistic warehouses~\cite{wurman2007robots, fragapane2021planning}.
In these situations, policy synthesis techniques, such as VI, must repeatedly update every state, resulting in severe runtime and memory overhead.
For example, open areas around $(0,0)$ may be relatively safe and uninteresting; thus, they can often be treated coarsely with little loss of accuracy.
Likewise, crowded corridors near shelves (W1--W3) can be prone to jams or delays; thus, a fine-grained analysis is required to ensure the robot's strategy remains robust.
Despite these semantic differences in the underlying system model, state-of-the-art PMC tools use VI and its variants~\cite{prism} and treat these regions the same, potentially wasting resources on trivial warehouse sections.

The key idea is not to treat all parts of the warehouse in the same way. \acronym\ refines uncertain or congested regions near W1--W3, while keeping simpler regions at a coarser level.

\section{\acronym}
\label{sec:approach}

\begin{figure}[t]
    \centering
    \includegraphics[width=\linewidth]{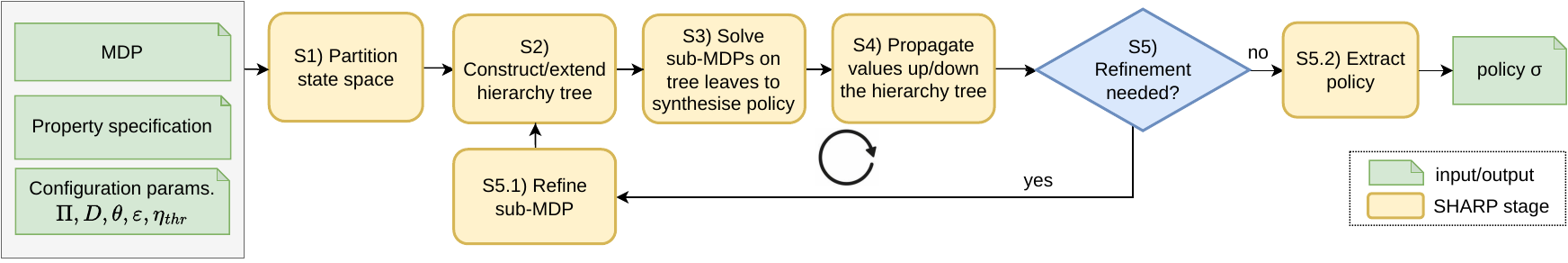}
    \caption{SHARP overview showing its various stages for hierarchical partitioning and adaptive refinement.}
    \Description{A workflow diagram showing the five stages of the SHARP algorithm: S1 Partitioning, S2 Tree Construction, S3 Local Solving, S4 Value Propagation, and S5 Refinement Check, with arrows indicating the process flow.}
    \label{fig:workflow}
\end{figure}

\acronym\ is a hierarchical partition-based approach for MDP policy synthesis that adaptively refines MDP partitions only where needed. 
Its high-level workflow is shown in Figure~\ref{fig:workflow} and comprises five stages (S1--S5). 
\acronym\ takes as inputs an MDP model, a reachability objective \emph{either}
$\PP_{\max}$ (maximise probability of reaching $G$) \emph{or}
$\RR_{\min}$ (minimise expected cumulative cost to $G$), a partition strategy $\Pi$, and four parameters: $D$ (maximum depth), $\theta$ (refinement threshold), $\varepsilon$~(solver tolerance), and $\eta_{\mathrm{thr}}$ (boundary-change threshold). The first two ($D$, $\theta$) control the hierarchical decomposition, while the latter two ($\varepsilon$, $\eta_{\mathrm{thr}}$) control the numerical accuracy and determine when changes in boundary values are significant enough to trigger re-solving, respectively.

The first \acronym\ stage (S1) involves executing an initial partitioning of the MDP into a set of \emph{blocks}, producing one sub-MDP per block.
\acronym\ uses this partitioning to construct the hierarchy tree (S2), whose root node corresponds to the entire set of states, and the children correspond to the individual blocks.
Each partition, at this stage, is considered a leaf node in the tree and is solved \textit{locally} using VI~(S3). This local solution uses fixed boundary conditions, derived from the global value vector $V$, to properly account for any transitions leaving the partition. This enables obtaining refined value estimates and a partial policy for each partition.

Once these refined estimates and partial policies are available, \acronym\ recursively propagates them up to their parent nodes in the tree (S4). Then, the iterative process is triggered (S3--S4) until values stabilise globally, re-solving leaves only when their boundary values change significantly.

  Then, \acronym\ checks whether further refinement is needed (S5). If this holds (S5.1), it refines the sub-MDP by subdividing selected
  partitions, extends the partition tree, and repeats the iterative process. Finally, if no further refinement is required and the global solution has converged, \acronym\ extracts the synthesised policy $\sigma$ (S5.2).
  
  During the execution of Algorithm~\ref{alg:SHARP-generic}, each invocation of \textsc{SolveLocally} produces both a value function and
  a local policy (the action achieving the Bellman optimum at each state).
  We then compose these local policies into a single global policy $\sigma$ over all states (Section~\ref{subsec:value-prop}).

\FloatBarrier

\subsection{Partitioning and Tree Hierarchy Construction}
\label{subsec:hierarchy-init} Foundational approaches to solving large MDPs include partitioning the state space to construct abstractions or smaller subproblems that can be solved more efficiently~\cite{dean1997min}. While in our work we do not perform state aggregation,
we leverage the principle of partitioning to structure the policy synthesis process hierarchically. \acronym\ is built upon two main ideas. First, the state space is partitioned into \emph{blocks}. Second, these blocks are organised hierarchically by recursively subdividing them, resulting in a nested structure represented by a \emph{hierarchy tree}.
\begin{definition}[Partitioning]\label{def:partitioning}
Given an MDP $\mathcal{M}$,
a \emph{partitioning} of its state space $S$ is a collection $\mathcal{B}=\{B_1, B_2, \dots, B_k\}$ of subsets of $S$, called \textbf{\emph{blocks}}, such that:
(i) Each block $B_i$ is non-empty ($B_i \neq \emptyset$ for all $i=1,\dots,k$); 
(ii) The blocks are pairwise disjoint ($B_i \cap B_j = \emptyset$ for all $i \neq j$);
and (iii) the union of all blocks covers the entire state space ($\bigcup_{i=1}^{k} B_i = S$).
\end{definition}

\begin{remark}[Symmetry]
The construction for probabilities extends verbatim to minimisation by replacing $\max$ with $\min$ in the Bellman operator. Our experiments focus on $\PP_{\max}$ and $\RR_{\min}$.
\end{remark}

\begin{definition}[Hierarchy tree]\label{def:tree}
A rooted tree $\mathcal{T}=(\mathcal{N},E)$ is a \emph{hierarchy tree} for $\mathcal{M}$ if 
(i) the root $r$ corresponds to the entire state space $S$;
(ii) every node $v\in \mathcal{N}$ corresponds to a block $B_v\subseteq S$; and
(iii) the blocks corresponding to the children of $v$ form a partitioning of $B_v$.
We denote the distance of $v$ from the root with $v.\mathrm{depth}$.
\end{definition}
Each node $v\in \mathcal{N}$ stores pointers to its parent and children, and also maintains its current local values ($v.\mathrm{values}$) and synthesised local policy ($v.\sigma$). To detect when re-solving is needed, leaf nodes additionally track the following metadata: (i) a snapshot of the boundary values from the previous solve ($v.\mathrm{prevBoundary}$); (ii) a flag indicating if the node has been solved at least once ($v.\mathrm{everSolved}$); and (iii) the residual error from the last local VI run ($v.\mathrm{localResidual}$).

\paragraph{Partition Strategy $\Pi$.}
We parameterise \acronym{} with a partition strategy $\Pi$ that provides: (i) $\Pi.\textsc{InitialPartition}(S)$; and (ii) $\Pi.\textsc{Refine}(B,d)$ at depth $d$.
Typical instantiations include uniform grid partitioning for MDPs with spatial state spaces and graph-based partitioning (e.g., counter-based, Strongly Connected Component (SCC) layering) for general MDPs.

\begin{algorithm}[t]
  \caption{\textsc{SHARP}: Scalable Hierarchical Adaptive Refinement for Policy Synthesis}
  \label{alg:SHARP-generic}
  \scriptsize
  \begin{algorithmic}[1]

  \Procedure{SHARP}{$\mathcal{M},\phi,G,\Pi,D,\theta,\varepsilon,\eta_{\mathrm{thr}}$}
\State \textbf{Init:} Create root on $S$. Initialise $V$ to $0$ on all states and fix goals:
$V(s){=}1$ for $s\in G$ under $\PP$, and $V(s){=}0$ for $s\in G$ under $\RR$.
    \State \textbf{Init pass:} $\mathcal{P}_0 \gets \Pi.\textsc{InitialPartition}(S)$; create children for all $B \in \mathcal{P}_0$
    \State \textbf{Solve/propagate:} \textsc{SolveLocally}(\emph{child},$\phi,\varepsilon$) for all children; \textsc{PropagateValues}(\texttt{root})
    \Repeat
      \State $V^{\text{old}} \gets V$ \Comment{Store for convergence check}
      \ForAll{leaves $v$ with \textsc{BoundaryChanged}$(v,\eta_{\mathrm{thr}})$}
        \State \textsc{SolveLocally}($v,\phi,\varepsilon$)
      \EndFor
      \State \textsc{PropagateValues}(\texttt{root})
      \State $L \gets \{\ell\text{ leaf} \mid \mathrm{depth}(\ell)<D \land \textsc{ShouldRefine}(\ell,\phi,\theta)\}$
      \ForAll{$\ell\in L$}
        \State \textsc{RefinePartition}($\ell,\Pi,\phi,\varepsilon$)
      \EndFor
      \State \textsc{PropagateValues}(\texttt{root})
    \Until{no leaf satisfies \textsc{BoundaryChanged}$(\cdot,\eta_{\mathrm{thr}})$, $L=\emptyset$, and $\|V - V^{\text{old}}\|_\infty \le \varepsilon$}
    \State \Return $(V,\sigma)$
  \EndProcedure

  \Procedure{SolveLocally}{$v,\phi,\varepsilon$} \Comment{Section~\ref{subsec:local-submdp}}
    \State Build the induced sub-MDP on $B_v$: add one-step outside successors as boundary states and fix their values from the global $V$
    \State Fix internal goals ($V{=}1$ for $\PP$, $V{=}0$ for $\RR$); preserve per-action costs $c(s,a)$
    \State Run VI (min/max per $\phi$) to tolerance $\varepsilon$; store $(V_v,\sigma_v)$; snapshot boundaries: $b_v^{\mathrm{prev}}$
  \EndProcedure

  \Procedure{RefinePartition}{$v,\Pi,\phi,\varepsilon$} \Comment{Section~\ref{subsec:adapt-refine}}
    \State Split $B_v$ via $\Pi$; create children; \textsc{SolveLocally}(\emph{child},$\phi,\varepsilon$) each
  \EndProcedure

  \Procedure{PropagateValues}{$v$} \Comment{Section~\ref{subsec:value-prop}}
    \State Post-order: recurse on all children of $v$ (if any), then copy each state's value and policy from its unique owning child
  \EndProcedure

  \Function{BoundaryChanged}{$v,\eta_{\mathrm{thr}}$}
    \State \Return if normalised boundary change exceeds $\eta_{\mathrm{thr}}$ \Comment{Section~\ref{subsec:local-submdp}}
  \EndFunction

  \Function{ShouldRefine}{$\ell,\phi,\theta$}
    \State $\theta_d \gets \textsc{AdaptThreshold}(\theta, \mathrm{depth}(\ell), \phi)$
    \State \Return $\tau(\ell) > \theta_d$ \Comment{Section~\ref{subsec:adapt-refine}}
  \EndFunction

  \end{algorithmic}
\end{algorithm}

\begin{figure*}[t]
\centering
\resizebox{0.99\textwidth}{!}{
\begin{tikzpicture}[
    >=latex,
    font={\scriptsize},
    scale=0.6, 
    transform shape,
    gridcell/.style={rectangle, draw, minimum size=0.7cm, fill=white},
    startcell/.style={rectangle, draw, minimum size=0.7cm, fill=green!30},
    goalcell/.style={rectangle, draw, minimum size=0.7cm, fill=blue!30},
    shelfcell/.style={rectangle, draw, minimum size=0.7cm, fill=black!60},
    uncertaincell/.style={rectangle, draw, minimum size=0.7cm, fill=red!20, opacity=0.2},
    boundarycell/.style={rectangle, draw, minimum size=0.7cm, fill=orange!15, dashed},
    block/.style={rectangle, draw, rounded corners=0pt, thick, draw=blue!70},
    zonehighlighted/.style={rectangle, draw, rounded corners=0pt, thick, draw=red!70},
    connector/.style={->, dashed, thick, blue},
    treelevel/.style={rectangle, draw, rounded corners, fill=gray!10, minimum height=1cm},
    note/.style={draw, rounded corners, fill=orange!10, align=left, text width=3.2cm, font={\small}},
    titlebox/.style={draw, rounded corners, fill=blue!5, align=center, text width=3.5cm}
]

\node[titlebox] at (-2, 6) {\textbf{Original MDP}\\(Warehouse Environment)};

\begin{scope}[shift={(-5,0)}]
    \foreach \x in {0,...,5} {
        \foreach \y in {0,...,5} {
            \node[gridcell] (cell-\x-\y) at (\x,\y) {};
            
            \node[font={\tiny}, gray!70] at (\x+0.15,\y+0.15) {\x,\y};
        }
    }
    
    \node[startcell] (cell-0-0) at (0,0) {};
    \begin{scope}[shift={(0,0)}]
        \draw[black] (0,0) node {
            \begin{tikzpicture}[scale=0.6]
                \fill[gray!80] (0,0) circle (0.2); 
                \fill[cyan!80] (-0.07,0.07) circle (0.05); 
                \fill[cyan!80] (0.07,0.07) circle (0.05);
                \draw[black, line width=0.6pt] (0,0.2) -- (0,0.3);
                \fill[red] (0,0.3) circle (0.04); 
            \end{tikzpicture}
        };
    \end{scope}
    \node[font={\tiny}, green!70!black] at (0.0,-0.25) {Start};
    
    \node[goalcell] (cell-5-5) at (5,5) {};
    \node[font={\tiny}, blue!70!black] at (5.0,5.25) {Goal};
    
    \foreach \y in {2,3} {
        \node[shelfcell] (cell-1-\y) at (1,\y) {};
    }
    \node[font={\tiny}, white] at (1,3) {W1};
    
    \foreach \y in {1,2} {
        \node[shelfcell] (cell-4-\y) at (4,\y) {};
    }
    \node[font={\tiny}, white] at (4,2) {W2};
    
    \foreach \x in {2,3} {
        \node[shelfcell] (cell-\x-5) at (\x,5) {};
    }
    \node[font={\tiny}, white] at (3,5) {W3};
    
    \node[block, fit=(cell-0-0)(cell-1-1)] (zone1) {};
    \node[block, fit=(cell-2-0)(cell-3-1)] (zone2) {};
    \node[block, fit=(cell-4-0)(cell-5-1)] (zone3) {};
    \node[block, fit=(cell-0-2)(cell-1-3)] (zone4) {};
    \node[block, fit=(cell-2-2)(cell-3-3)] (zone5) {};
    \node[block, fit=(cell-4-2)(cell-5-3)] (zone6) {};
    \node[block, fit=(cell-0-4)(cell-1-5)] (zone7) {};
    \node[block, fit=(cell-2-4)(cell-3-5)] (zone8) {};
    \node[zonehighlighted, fit=(cell-4-4)(cell-5-5)] (zone9) {};
    
    \node[blue, font={\tiny}] at (0.5,0.5) {Block 1};
    \node[blue, font={\tiny}] at (2.5,0.5) {Block 2};
    \node[blue, font={\tiny}] at (4.5,0.5) {Block 3};
    \node[blue, font={\tiny}] at (0.5,2.5) {Block 4};
    \node[blue, font={\tiny}] at (2.5,2.5) {Block 5};
    \node[blue, font={\tiny}] at (4.5,2.5) {Block 6};
    \node[blue, font={\tiny}] at (0.5,4.5) {Block 7};
    \node[blue, font={\tiny}] at (2.5,4.5) {Block 8};
    \node[red, font={\tiny}] at (4.5,4.5) {Block 9};

    \node[red!70!black, font={\tiny}, rounded corners=1pt, inner sep=0.5pt] at (4.85,3.85) {B9a};
    \node[red!70!black, font={\tiny}, rounded corners=1pt, inner sep=0.5pt] at (3.85,4.85) {B9b};
    \node[red!70!black, font={\tiny}, rounded corners=1pt, inner sep=0.5pt] at (3.85,3.85) {B9c};
    \node[red!70!black, font={\tiny}, rounded corners=1pt, inner sep=0.5pt] at (4.85,4.85) {B9d};
    
    \draw[thick, orange, dashed] (cell-2-0) circle (0.25cm); 
    \draw[thick, orange, dashed] (cell-2-1) circle (0.25cm); 
    \draw[thick, orange, dashed] (cell-0-2) circle (0.25cm); 
    
    \node[orange, font={\tiny}] at (2.2, 0.2) {$b_1$};
    \node[orange, font={\tiny}] at (2.2, 1.2) {$b_2$};
    \node[orange, font={\tiny}] at (0.2, 2.2) {$b_3$};
\end{scope}

\node[titlebox] at (5, 6) {\textbf{Hierarchy }};

\node[treelevel, minimum width=4cm] (rootlevel) at (5, 4.2) {};

\node[treelevel, minimum width=10cm, xshift=1cm] (zonelevel) at (5, 2.4) {};
\node[treelevel, minimum width=6cm] (refinelevel) at (9.9, 1) {};

\node[circle, draw, fill=yellow!10, minimum size=0.8cm] (root) at (5, 4.2) {$S$};

\node[circle, draw, fill=green!10, minimum size=0.6cm] (n1) at (2.0, 2.4) {B1};
\node[circle, draw, fill=green!10, minimum size=0.6cm] (n2) at (3.0, 2.4) {B2};
\node[circle, draw, fill=green!10, minimum size=0.6cm] (n3) at (4.0, 2.4) {B3};
\node[circle, draw, fill=green!10, minimum size=0.6cm] (n4) at (5.0, 2.4) {B4};
\node[circle, draw, fill=green!10, minimum size=0.6cm] (n5) at (6.0, 2.4) {B5};
\node[circle, draw, fill=green!10, minimum size=0.6cm] (n6) at (7.0, 2.4) {B6};
\node[circle, draw, fill=green!10, minimum size=0.6cm] (n7) at (8.0, 2.4) {B7};
\node[circle, draw, fill=green!10, minimum size=0.6cm] (n8) at (9.0, 2.4) {B8};
\node[circle, draw=red!70, thick, fill=red!10, minimum size=0.6cm] (n9) at (10.0, 2.4) {B9};

\node[circle, draw, fill=green!10, minimum size=0.6cm] (n9a) at (8.9, 1) {B9a};
\node[circle, draw, fill=green!10, minimum size=0.6cm] (n9b) at (9.9, 1) {B9b};
\node[circle, draw, fill=green!10, minimum size=0.6cm] (n9c) at (10.9, 1) {B9c};
\node[circle, draw=blue!70, thick, fill=blue!10, minimum size=0.6cm] (n9d) at (11.9, 1) {B9d};

\draw[->] (root) -- (n1);
\draw[->] (root) -- (n2);
\draw[->] (root) -- (n3);
\draw[->] (root) -- (n4);
\draw[->] (root) -- (n5);
\draw[->] (root) -- (n6);
\draw[->] (root) -- (n7);
\draw[->] (root) -- (n8);
\draw[->] (root) -- (n9);

\draw[->] (n9) -- (n9a);
\draw[->] (n9) -- (n9b);
\draw[->] (n9) -- (n9c);
\draw[->] (n9) -- (n9d);

\node[titlebox] at (10, -2.2) {\textbf{Example: Sub-MDP for Block 1}};
\begin{scope}[shift={(5,-3.5)}]
    \node[startcell] (sub-0-0) at (0,0) {};
    \node[anchor=south] at (0, -0.05) {\textbf{Start}};
    
    \begin{scope}[shift={(0,0)}]
        \draw[black] (0,0) node {
            \begin{tikzpicture}[scale=0.6]
                \fill[gray!80] (0,0) circle (0.2); 
                \fill[cyan!80] (-0.07,0.07) circle (0.05);
                \fill[cyan!80] (0.07,0.07) circle (0.05);
                \draw[black, line width=0.6pt] (0,0.2) -- (0,0.3);
                \fill[red] (0,0.3) circle (0.04);
            \end{tikzpicture}
        };
    \end{scope}
    
    \node[gridcell] (sub-1-0) at (1,0) {};
    \node[gridcell] (sub-0-1) at (0,1) {};
    \node[gridcell] (sub-1-1) at (1,1) {};
    
    \node[boundarycell] (bound-20) at (2,0) {}; 
    \node[boundarycell] (bound-21) at (2,1) {}; 
    \node[boundarycell] (bound-02) at (0,2) {}; 

    \node[block, fit=(sub-0-0)(sub-1-1)] (sub-block) {};
    \node[blue] at (0.75,0.75) {\large Block 1};
    
    \node[font=\tiny, gray!70] at (0.15,0.15) {0,0};
    \node[font=\tiny, gray!70] at (1.15,0.15) {1,0};
    \node[font=\tiny, gray!70] at (0.15,1.15) {0,1};
    \node[font=\tiny, gray!70] at (1.15,1.15) {1,1};
    
    \node[font=\tiny, orange] at (2.15,0.15) {2,0};
    \node[font=\tiny, orange] at (2.15,1.15) {2,1};
    \node[font=\tiny, orange] at (0.15,2.15) {0,2};
    
    \node[orange, font=\tiny] at (2.5, 0.5) {from B2};
    \node[orange, font=\tiny] at (0.5, 2.5) {from B4};
\end{scope}

\draw[->, thick, orange!90!black] (sub-block) to[out=135, in=270]
  node[below left, align=center, text width=2.2cm]{Store local solution in B1} (n1);

\draw[->, thick, red!70!black] (n1) to[out=135, in=225]
  node[left, align=left, text width=1.8cm]{Propagate up} (root);

\node[draw, rounded corners, fill=white, text width=4.5cm, align=left, font=\small] at (-3.1,-3) {
    \textbf{Legend:}
    \begin{itemize}\itemsep=0.3cm
        \item \tikz\node[draw=red!70,thick,fill=red!10,circle,inner sep=3pt]{};\ Refined block
        \item \tikz\draw[thick, blue] (0,0) rectangle (0.5,0.25);\ Block boundary
        \item \tikz\draw[thick, orange, dashed] (0.25,0.125) circle (0.125cm);\ Boundary state
    \end{itemize}
};

\draw[->, thick, orange] (n2) to[out=270, in=45] (bound-20) to[out=180, in=0] (sub-1-0);
\draw[->, thick, orange] (n2) to[out=270, in=45] (bound-21) to[out=180, in=0] (sub-1-1);
\draw[->, thick, orange] (n4) to[out=270, in=135] (bound-02) to[out=270, in=90] (sub-0-1);

\node[note] at (16.0, 5.2) {
  \textbf{Stages 1-2:} Partition state space and construct hierarchy tree with root $S$. Block~9 marked for refinement.
};
\node[note] at (16.0, 2.2) {
  \textbf{Stage 3:} Solve sub-MDPs on tree leaves with boundary states from neighbours.
};
\node[note] at (16.0, -0.2) {
  \textbf{Stage 4:} Propagate values up the hierarchy tree
};

\node[note] at (16.0, -3.75) {
  \textbf{Stage 5:} Check if refinement needed
  \begin{itemize}
    \item S5.1: Refine sub-MDP (if yes)
    \item S5.2: Extract policy (if no)
  \end{itemize}
};

\end{tikzpicture}
}
\caption{\textbf{Overview of Sub-MDP Creation and Hierarchical Refinement in \acronym.}
A $6\times6$ warehouse with shelves (W1--W3) is partitioned into nine blocks.
\emph{Block~9} contains the goal and is refined further.
\emph{Block~1} contains the start state, and its dashed boundary states connect to adjacent \emph{Block~2} and \emph{Block~4}.
Solved values then propagate back to the parent node.}
\Description{A composite diagram showing a 6 by 6 warehouse MDP, its hierarchy tree, and a local sub-MDP for Block 1. The figure illustrates how the original warehouse is partitioned into blocks, how Block 9 is refined, how boundary states are introduced for a local solve, and how values are propagated back up the hierarchy.}
\label{fig:submdp-example}
\end{figure*}

Algorithm~\ref{alg:SHARP-generic} summarises the \acronym\ workflow introduced above. 
It takes the inputs described in Section~\ref{sec:approach} and returns a value vector $V$ for the MDP states $S$ and a memoryless policy~$\sigma$. 
We first initialise the global value vector to $0$ for all non-goal states.
The values of the goal states are then fixed according to the type of the objective.
Specifically, we set $V(s){=}1$ for probability objectives and $V(s){=}0$
for reward objectives.
These assignments provide the boundary conditions that are used when
the initial blocks are solved locally.
\noindent
\\
\textbf{Example 1.}
Consider again the warehouse robotics model from Section~\ref{sec:motiv-example}.
Our goal is to synthesise an optimal policy for the reachability property
\(\PP_{\max}\!\bigl[\,\mathsf{F}\,\text{``goal''}\bigr]\).

\noindent
\emph{Hierarchy initialisation:} 
We begin with a single root node that contains all states of the MDP, that is the 36 grid positions of the 6$\times$6 grid.
Next, we apply a coarse geometric partition to the grid: a uniform 3$\times$3 tiling yields nine non-overlapping blocks (Figure~\ref{fig:submdp-example}).
Each block is non-empty, and together they cover the entire warehouse, satisfying the conditions of Definition~\ref{def:partitioning}; each becomes a child of the root node, forming a hierarchy tree (Definition~\ref{def:tree}) with a proper partition of the state space.

After the initial partitioning, \acronym{} solves each of the nine blocks locally using \textsc{SolveLocally} with boundary values from $V$, then propagates values up to the root. The algorithm then enters its main loop, where it may refine blocks whose value spread exceeds $\theta$.
For instance, block 9 containing the goal area might be refined further if its value spread exceeds $\theta$.

\subsection{Adaptive Partition Refinement}
\label{subsec:adapt-refine}
In order to decide whether to refine a leaf node or not in the hierarchy tree,
we create a refinement criterion that compares the value spread $\Delta(v)$ within each block against a threshold.

\textbf{Refinement criterion.}
For a node $v$ with block $B_v$, we define the value spread as $\Delta(v) = \max_{s\in B_v}V_v(s) - \min_{s\in B_v}V_v(s)$.
If this spread is small, then the block is already fairly homogeneous and there is little reason to split it further. If, on the other hand, the spread is large, then the block is mixing states with rather different values, which can affect the resulting policy.
For \(\PP_{\max}\) we simply use \(\tau(v) = \Delta(v)\).
For \(\RR_{\min}\), we use either the same absolute spread in the grid-based warehouse experiments or the normalised spread \(\Delta(v)\,/\,(|\mathrm{avg}(V_v)|+\beta)\) in the non-grid experiments.
A leaf $v$ at depth $d < D$ is refined once $\tau(v)$ exceeds the depth-adapted threshold $\theta_d$.\footnote{\textsc{AdaptThreshold} computes $\theta_d$ internally. Concrete settings for each instantiation appear in Section~\ref{sec:evaluation}.}
  This is the test implemented by \textsc{ShouldRefine}$(v,\phi,\theta)$ in Algorithm~\ref{alg:SHARP-generic}.
  When the test succeeds, \textsc{RefinePartition} splits the block into disjoint child blocks, solves them using boundary values from $V$, and propagates the updated values upward.

\noindent
\\
\textbf{Example 2 (continued).}
Continuing the warehouse scenario of Example 1, suppose that the nine
top-level partitions in Figure~\ref{fig:submdp-example} have already been solved,
and that partition~9, which contains the goal cell \((5,5)\), shows a large value
spread~\(\Delta\).
This is not surprising, since states close to the goal have values close to \(1.0\),
whereas states one step farther away are noticeably smaller, and so
\(\Delta>\theta\). The block is therefore refined. This refinement of partition~9 separates its four cells into the child blocks \texttt{B9a}--\texttt{B9d}, isolating the goal cell and nearby high-value states from the lower-value states farther away.

\subsection{Local Sub-MDP Policy Synthesis}
\label{subsec:local-submdp}
Once a node $v$ is partitioned, either as part of the initial partitioning or due to further refinement, \acronym\ views each partition as a sub-MDP that includes both the partition's states and boundary states. 
The next definitions formalise this local-solving step. In particular, we generate the induced sub-MDP of a block $B$ by adding its one-step outside successors as boundary states (Def.~\ref{def:submdp}), making them absorbing, and assigning them fixed values from the current global vector $V$ (Def.~\ref{def:boundary_value}); local VI then solves the standard Bellman equations inside this closed sub-MDP.

\textbf{Constructing the sub-MDP.}
\acronym\ builds the induced sub-MDP $\mathcal{M}\!_B$ for block $B\subseteq S$ as specified in Definition~\ref{def:submdp} below. Boundary states $\mathrm{bd}(B)$ are made absorbing, and their values are fixed from the current global vector $V$ as specified in Definition~\ref{def:boundary_value}.

\begin{definition}[Induced sub-MDP] 
\label{def:submdp}
Let $\mathcal{M}=(S,\bar s,A,\delta,c,AP,L)$ be an MDP and $B\subseteq S$, $B\neq\emptyset$.
The boundary is $\mathrm{bd}(B) = \{t\in S\setminus B \mid \exists s \in B, a\in A(s): \delta(s,a)(t)>0\}$.
The \emph{sub-MDP induced by $B$} is $\mathcal{M}_B = (B\cup\mathrm{bd}(B), \bar s_B, A_B, \delta_B, c_B, AP, L_B)$
with $A_B = A$, $c_B = c$, $L_B = L$, and for $s\in B\cup\mathrm{bd}(B)$, $a\in A(s)$:
\[
\bar s_B = \begin{cases}
\bar s & \text{if }\bar s\in B\\
\text{any in }B & \text{otherwise}
\end{cases}, \quad
\delta_B(s,a)(t) = \begin{cases}
\delta(s,a)(t) & s\in B, t\in B\cup\mathrm{bd}(B)\\
1 & s\in\mathrm{bd}(B), t=s\\
0 & \text{otherwise}
\end{cases}
\]
\end{definition}

Accordingly, each state in $\mathrm{bd}(B)$ becomes absorbing in the sub-MDP
$\mathcal{M}\!_B$. 

The following definition describes how the local solve fixes its value using $V(t)$ as a fixed boundary condition, i.e., whenever a transition from $B$ reaches a boundary state $t$, we use the current global estimate $V(t)$.

\begin{definition}[Boundary-value assignment]\label{def:boundary_value}
Let $B\subseteq S$ be a block represented by a node $v$ in the hierarchy
tree~$\mathcal{T}$, and let
$\mathcal{M}\!_B$ be the sub-MDP induced by $B$
with boundary states $\mathrm{bd}(B)$.
For every boundary state $t\in\mathrm{bd}(B)$, its \emph{boundary value}
is set to $V_B(t)\;=\;V(t)$,
where $V$ is the
global value vector maintained at the
root of~$\mathcal{T}$.
These values are held fixed throughout the local VI step on
$\mathcal{M}\!_B$, thereby incorporating the influence of
the surrounding MDP.
That is, the Bellman equations are solved only for internal non-goal states in $B$; boundary values are held fixed and excluded from Bellman updates.
\end{definition}

\paragraph{Boundary Change Detection.}\label{par:boundary-change}
During the iterative refinement process, SHARP must detect when a node's boundary values have
changed significantly enough to trigger the re-solving process of its local sub-MDP.
Let $\mathrm{bd}(B_v)$ denote the boundary states of block $B_v$,
and let $b_v^{\mathrm{prev}} := V^{\text{(last solve)}}|_{\mathrm{bd}(B_v)}$ be the boundary snapshot stored after $v$'s previous local solve.
We define the \textsc{BoundaryChanged} function as:

\[
\textsc{BoundaryChanged}(v,\eta_{\mathrm{thr}}) \,:=\, \frac{\bigl\|\,V|_{\mathrm{bd}(B_v)} - b_v^{\mathrm{prev}}\,\bigr\|_\infty}
     {\max\bigl\{1,\,\bigl\|\,b_v^{\mathrm{prev}}\,\bigr\|_\infty\bigr\}} \,>\, \eta_{\mathrm{thr}}.
\]

When this predicate evaluates to false (i.e., the boundary change is not significant enough to trigger re-solving), it implies the absolute mismatch bound
$\bigl\|\,V|_{\mathrm{bd}(B_v)} - b_v^{\mathrm{prev}}\,\bigr\|_\infty \le \eta_{\mathrm{thr}}\cdot \max\{1,\|b_v^{\mathrm{prev}}\|_\infty\}$. We use this absolute quantity in the residual bounds.
Newly created leaves are solved immediately upon creation (\textsc{RefinePartition}), establishing $b_v^{\mathrm{prev}}$ before the predicate is ever evaluated.

\textbf{Local VI.}
The sub-MDP is solved using VI until
the local Bellman residual $\lVert T_B V - V\rVert_\infty<\varepsilon$,
where $T_B$ is the local Bellman operator on $\mathcal{M}_B$
(equivalently, for synchronous VI, $\lVert V^{(k+1)}-V^{(k)}\rVert_\infty<\varepsilon$).
The action that attains the optimum value is stored as the local policy.

\begin{definition}[Local value function]\label{def:local_value_a}
Given a reachability property $\PP_{\max}[\mathsf{F}\,G]$ or $\RR_{\min}[\mathsf{F}\,G]$ with global goal set $G$,
fix a block $B$ and its induced sub-MDP
$\mathcal M\!_B$ (Def.~\ref{def:submdp}).
Let $G_B=G\cap B$ and use $\mathrm{bd}(B)$ for the boundary states.

The \emph{local value} $V_B:B\cup\mathrm{bd}(B)\to\mathbb R$ is the unique solution of:

\noindent\textbf{Probability objective} ($\PP_{\max}$)
\[
  V_B(s)=
  \begin{cases}
    1, & s\in G_B,\\
    \displaystyle
    \max_{a\in A_B(s)}
      \sum_{t\in B\cup\mathrm{bd}(B)}
        \delta_B(s,a)(t)\,V_B(t),
      & s\in B\setminus G_B,\\
    V(s), & s\in\mathrm{bd}(B).
  \end{cases}
\]

\noindent\textbf{Cost objective} ($\RR_{\min}$)
\[
  V_B(s)=
  \begin{cases}
    0, & s\in G_B,\\
    \displaystyle
    \min_{a\in A_B(s)}
      \Bigl( c(s,a) + \sum_{t\in B\cup\mathrm{bd}(B)}
        \delta_B(s,a)(t)\,V_B(t) \Bigr), & s\in B\setminus G_B,\\
    V(s), & s\in\mathrm{bd}(B).
  \end{cases}
\]

\end{definition}

\textbf{Storing and propagating.}
After local convergence, \acronym\ stores both the value vector $\{V_B(s) \mid s\in B \cup \mathrm{bd}(B)\}$ and the corresponding partial policy $\sigma$ in the
node associated with block~$B$.
If the node is a leaf, the sub-MDP solving is complete.
Otherwise, once all child nodes are solved, we propagate their values upward to update this parent node's estimates as
detailed in Section~\ref{subsec:value-prop}.

\noindent
\\
\textbf{Example 3 (continued).} 
Suppose we refine B9 into four single-state child blocks: \texttt{B9a}, \texttt{B9b}, \texttt{B9c}, and \texttt{B9d},
containing states $(5,4)$, $(4,5)$, $(4,4)$, and the goal $(5,5)$, respectively.
The goal child \texttt{B9d} has its value fixed at $1$ and, for this reason, requires no solving.
In practice, the algorithm would likely stop refinement before reaching single-state blocks, but we use those here in order to show how our approach works.
Following Definition~\ref{def:submdp}, to construct the sub-MDP for \texttt{B9a} (containing cell $(5,4)$), we include: (i) the state $(5,4)$ with all its original actions and transitions,
and (ii) any states directly reachable from $(5,4)$ and lying outside \texttt{B9a} as boundary states $\mathrm{bd}(\texttt{B9a})$, including the goal $(5,5)$ and adjacent cells.
These boundary states are absorbing in the sub-MDP (self-loops with probability 1) and have been assigned fixed values from $V$ (Definition~\ref{def:boundary_value}).
For such a small sub-MDP, local VI converges quickly. Since $(5,4)$ is adjacent to the goal (whose value is 1.0), the converged value for $(5,4)$ will be high.
Moreover, the local solve also produces a policy: $\sigma_{\texttt{B9a}}(5,4) = \texttt{up}$, since moving up toward the goal $(5,5)$ maximises the reachability probability.
After solving the three non-goal child sub-MDPs, we propagate their values and policies back to update block B9's estimates.

\subsection{Value Propagation Up the Tree Hierarchy}
\label{subsec:value-prop}
Having described how local sub-MDPs are solved (Section~\ref{subsec:local-submdp}), we now describe how their solutions are combined to produce a globally consistent policy.

\paragraph{Global Policy Composition.}

Since leaf nodes partition $S$, each state $s$ belongs to exactly one leaf $v$,
which stores a local value function $V_v$ and a deterministic memoryless policy $\sigma_v : B_v \to A$ attaining the Bellman optimum (\textsc{SolveLocally}).
For each $s \in S$, let $\mathrm{leaf}(s)$ denote the unique leaf with $s \in B_{\mathrm{leaf}(s)}$.
We can therefore define the global policy by simply copying from the leaves:
$\sigma(s) := \sigma_{\mathrm{leaf}(s)}(s)$ and $V(s) := V_{\mathrm{leaf}(s)}(s)$, which essentially is the action/value that the owning leaf deemed optimal.
This composition is well-defined because the partition is disjoint and no state appears in more than one leaf.
\textsc{PropagateValues} implements this copying as a bottom-up traversal of $\mathcal{T}$, i.e., 
for each node $v$, it recurses into children first, then pulls their values/policies into the parent.
Since children partition $B_v$, \acronym\ (Algorithm~\ref{alg:SHARP-generic}) by construction ensures that no two children claim the same state, meaning that the root ends up with a globally consistent policy $\sigma$.
We formally show in Section~\ref{subsec:sharp-theory} that the policy thus composed is near-optimal when boundary mismatch is kept small.

\textbf{Example 4 (continued).}
Following the \textsc{PropagateValues} procedure of Algorithm~\ref{alg:SHARP-generic}, with the local solutions for \texttt{B9a}, \texttt{B9b}, and \texttt{B9c} now computed, we merge them back into block B9 by assigning each sub-block's value (and policy) to the corresponding cell.
For instance, if \texttt{B9a} determined that $(5,4)$ has a probability $\approx\! 1.0$ to reach $(5,5)$, we set $V_9(5,4) \approx 1.0$ and copy the policy $\sigma_9(5,4) = \texttt{up}$ from $\sigma_{\texttt{B9a}}$. Similarly, \texttt{B9b} and \texttt{B9c} provide values and policies for their respective cells. 
At this point, block B9's value estimates are updated based on its children's refined solutions. 
Although in this example, we reached single-state blocks for illustration, \acronym, in general, checks in subsequent iterations if further refinement is needed based on the updated value spread. 

\subsection{Theoretical Guarantees}
\label{subsec:sharp-theory}
We show that, under the assumptions stated below, \acronym\ returns a near-optimal final value vector when each leaf sub-MDP is solved to tolerance $\varepsilon$ and the boundary mismatch at termination remains small.
Let $\eta$ denote the maximum boundary mismatch at termination, i.e., the largest absolute difference between the boundary values used in a leaf's last local solve and the final global values (defined formally below).
The \emph{global Bellman residual} then satisfies $\|T V^{\mathrm{fin}} - V^{\mathrm{fin}}\|_\infty \le \varepsilon+\eta$.
Furthermore, under the assumptions stated below, this residual bound allows us to obtain a formal guarantee on the actual error by standard fixed-point arguments~\cite{denardo1967dp,bertsekas1996neuro}.
We analyse the two objectives used in our evaluation: minimum expected cost ($\RR_{\min}$) and maximum reachability probability ($\PP_{\max}$). 
Our analysis focuses on the main arguments; complete proofs are provided in the supplementary material.

We first state the assumptions used below, which follow the standard SSP and reachability assumptions for MDPs~\cite{baier2008principles}:
  \begin{enumerate}[topsep=4pt,itemsep=2pt]
  \item \textbf{$\RR_{\min}$:} The MDP is an SSP instance with the standard conditions, as defined
  in Section~\ref{sec:prelims}:
  (i) a proper policy exists; (ii) costs are non-negative with zero cost at goals; (iii) goals are absorbing; and (iv) every 
  improper policy has infinite expected cost from some state~\cite{bertsekas1991ssp,bertsekas1996neuro}.
  \item \textbf{$\PP_{\max}$:} \emph{Uniform absorption:} there exists an $\alpha>0$ such that from every non-absorbing state, under any action, the probability to enter the absorbing set (goal or sink) in one step is at least $\alpha$~\cite{puterman1994mdps}.
  \end{enumerate}

In our case, all $\RR_{\min}$ instances satisfy the SSP assumptions. For the $\PP_{\max}$ warehouse benchmarks, the relevant assumption is the uniform-absorption condition used in part~(b), rather than SSP. More precisely, from every non-absorbing state, every enabled action reaches the absorbing set, that is, either the goal or the sink, in one step with probability at least $\alpha>0$. In our models, $\alpha$ is given by \texttt{P\_FAIL}. Furthermore, both the goal and sink states are absorbing, with values fixed to $V(s)=1$ for goal states and $V(s)=0$ for sink states.

\noindent\emph{Boundary changes and re-solving strategy.} Let $T$ be the Bellman optimality operator with fixed point~$V^*$.
During the execution of the algorithm, each leaf sub-MDP is solved to a specified tolerance~$\varepsilon$ using the \textsc{SolveLocally} procedure,
with fixed boundary values taken from the current global vector at the time of solving.
Moreover, note that blocks are not re-solved immediately when boundary values change; instead,
we take a snapshot of these boundary values at each solve (as specified in \textsc{SolveLocally})
and only trigger re-solving when \textsc{BoundaryChanged} detects that the change exceeds
a threshold at runtime.
This yields an \emph{asynchronous fixed-point} schedule~\cite{bertsekas1983dist}; our contribution is the hierarchical partitioning and boundary-change trigger that decides where and when to re-solve.
 However, this also means that changes below this threshold will
remain unresolved, and they continue to accumulate until the algorithm's termination.
This maximum accumulated discrepancy is captured by $\eta$, measured as $\eta \;=\; \max_{\text{leaf } B}\; \max_{t \in \mathrm{bd}(B)} \bigl|\,V^{\mathrm{fin}}(t) - V^{\mathrm{used}}_B(t)\,\bigr|$, where $V^{\mathrm{fin}}$ is the final global value vector when SHARP terminates and $V^{\mathrm{used}}_B(t)$
denotes the boundary value that block $B$ used at its most recent solve.

\noindent\emph{On the re-solve trigger.} In the implementation, we do not check $\eta$ directly. Instead, we use the relative test \textsc{BoundaryChanged}$(\cdot,\eta_{\mathrm{thr}})$ defined above.
When \acronym{} terminates, this test is false for every leaf. This gives the absolute mismatch bound
$\bigl\|V|_{\mathrm{bd}(B)} - b^{\mathrm{prev}}\bigr\|_{\infty} \le \eta_{\mathrm{thr}}\cdot\max\{1, \|b^{\mathrm{prev}}\|_{\infty}\}$,
and therefore $\eta$ is bounded in terms of $\eta_{\mathrm{thr}}$ (scaled by the chosen normalisation), allowing the error bounds in Theorem~\ref{thm:main} to be stated directly in terms of the input threshold. The next result shows how the residual bound translates into a bound on the actual error.

\begin{theorem}[Convergence and Error]
\label{thm:main}
When \acronym{} terminates, its final value vector $V^{\mathrm{fin}}$ is a close approximation of the true optimal solution $V^*$. First, the final solution is nearly stable, in the sense that the global Bellman residual is bounded by the local solver tolerance and the measured boundary mismatch: $\|T V^{\mathrm{fin}} - V^{\mathrm{fin}}\|_{\infty} \le \varepsilon + \eta$.
This residual bound, in turn, yields bounds on the true error $\|V^{\mathrm{fin}} - V^*\|_\infty$:
\begin{enumerate}[topsep=4pt,itemsep=2pt]
\item[(a)] \textbf{Minimum expected cost ($\RR_{\min}$).} For SSP problems, $\|V^{\mathrm{fin}} - V^*\|_{\infty} \le C\,(\varepsilon + \eta)$, where $C$ depends on the expected number of steps to reach the goal under suitable proper policies. In other words, the error scales with $\varepsilon + \eta$, up to the constant $C$.
\item[(b)] \textbf{Maximum reachability probability ($\PP_{\max}$).} For $\PP_{\max}$, we additionally assume that from any non-absorbing state, every action available in that state reaches either the goal or the sink in one step with probability at least $\alpha>0$. Under this assumption, $\|V^{\mathrm{fin}} - V^*\|_{\infty} \le (\varepsilon + \eta)/\alpha$.
\end{enumerate}
\end{theorem}

Intuitively, if each leaf is solved accurately (within $\varepsilon$) and boundary changes that remain at stopping are small (captured by $\eta$), then the assembled solution is near-optimal.
The theorem above concerns the final value vector. We next relate this guarantee to the global policy obtained by composing the local leaf policies.

\paragraph{Policy Quality Guarantee.}

Each leaf policy $\sigma_v$ is optimal with respect to the local value $V_v$. Boundary values may drift by up to $\eta$ between the local solve and termination of the algorithm.
Since transition probabilities sum to one, a one-step Bellman backup under the composed policy $\sigma$ can therefore change by at most $\eta$ when the boundary values change.
Combining this with the local solver tolerance $\varepsilon$, the Bellman residual under $\sigma$ is bounded by $\varepsilon{+}\eta$.
Based upon standard contraction mapping arguments~\cite{bertsekas1996neuro,denardo1967dp} and the global error bound $(\varepsilon{+}\eta)/\alpha$ from Theorem~\ref{thm:main},
we obtain $\|V^\sigma - V^*\|_\infty \le (2\varepsilon+2\eta)/\alpha$ for $\PP_{\max}$ under uniform absorption.
For $\RR_{\min}$, an analogous bound can be obtained when the composed policy $\sigma$ is proper.

\acronym{} terminates in finitely many iterations, as we now state formally.
\begin{theorem}[Termination]
\label{thm:termination}
Given bounded refinement depth $D$ and fixed thresholds $\varepsilon,\eta_{\mathrm{thr}}>0$, if in each outer pass every leaf satisfying \textsc{BoundaryChanged} is re-solved via \textsc{SolveLocally},
then \acronym\ terminates in finite time.
\end{theorem}

\section{Implementation}
\label{sec:implementation}
\acronym\ is implemented in Java on top of the probabilistic model checker PRISM~\cite{prism}. 
The core of our implementation lies in newly developed algorithms integrated within PRISM's explicit engine, including hierarchical partitioning, adaptive refinement based on value spread, sub-MDP construction with boundary states, and bottom-up value propagation.
\acronym\ can be configured by selecting a partition strategy $\Pi$ (grids: $(N_x,N_y)$; general MDPs: SCC-based or counter-based), as well as the maximum refinement depth $D$, refinement threshold $\theta$, solver tolerance $\varepsilon$, and boundary-change threshold $\eta_{\mathrm{thr}}$.
The \acronym\ implementation, case studies, and experimental results are available at \url{https://github.com/alexEvangelidis/sharp}.

\section{Evaluation}
\label{sec:evaluation}

\subsection{Research Questions}
We use the research questions below to assess \acronym\ and extract decision-making insights,
helping practitioners effectively embed \acronym\ in their verification workflows.

\noindent\textbf{RQ1 (Efficiency):} 
How efficient is \acronym\ compared to PRISM in terms of execution time and memory overhead across various problem instances?

\noindent\textbf{RQ2 (Scalability):} 
How does \acronym\ scale
with problem instances of increasing sizes in terms of states and transitions?

\noindent\textbf{RQ3 (Generalisation):} 
How well does \acronym\ generalise to different MDP structures beyond spatial models, and what structural properties determine its effectiveness?

\noindent\textbf{RQ4 (Hyperparameter sensitivity):} 
How is \acronym's performance affected by adjusting hyperparameters: initial partition size, maximum depth, and refinement thresholds?

\subsection{Experimental Setup}
\label{sec:exp_setup}
\noindent
\textbf{Case studies.}
For the evaluation, we use two benchmark groups. The first is a set of warehouse MDPs (Table~\ref{tab:case_studies}), and the second is a set of non-grid benchmarks drawn from RL~\cite{kaelbling1996rlsurvey,sutton1999temporal,deepmind2017gridworlds} and verification~\cite{velasquez2022controller,bals2024multigain,cai2021LTL}. For the warehouse models, we consider $512\times512$ and $1024\times1024$ grids with three obstacle layouts, namely NW, SW, and MW. For each layout, we study two variants. In the Pmax case, the objective is to maximise the probability of reaching the goal in the presence of a failure sink, with \texttt{P\_FAIL}=$5\times 10^{-4}$ and \texttt{P\_SUCC}\,$\in\{0.80,0.90\}$. In the Rmin case, the objective is to minimise the expected number of steps, assuming 0.8 probability of moving as intended, 0.2 probability of a self-loop, and unit step cost.
Additionally, we tested 5 random $1024\times1024$ MW MDPs with varied wall configurations. 

\begin{table}[!t]
\centering
\caption{MDP warehouse benchmarks. NW: no walls, SW: single wall, MW: multiple walls.}
\label{tab:case_studies}
\setlength{\tabcolsep}{3.5pt}
\small
\begin{tabular}{l c c c c l l}
\toprule
\textbf{Grid} & \textbf{\#States} & \textbf{Walls} & \textbf{\#Trans} & \textbf{\#Trans} & \textbf{ID} & \textbf{ID} \\
& & & \textbf{($\PP_{\max}$)} & \textbf{($\RR_{\min}$)} & \textbf{($\PP_{\max}$)} & \textbf{($\RR_{\min}$)} \\
\midrule
\multirow{3}{*}{512$^2$} & \multirow{3}{*}{262K} & NW & 3.14M & 2.10M & W512NW-a & W512NW-b \\
& & SW & 3.14M & 2.09M & W512SW-a & W512SW-b \\
& & MW & 3.14M & 2.09M & W512MW-a & W512MW-b \\
\midrule
\multirow{3}{*}{1024$^2$} & \multirow{3}{*}{1.05M} & NW & 12.58M & 8.38M & W1024NW-a & W1024NW-b \\
& & SW & 12.57M & 8.38M & W1024SW-a & W1024SW-b \\
& & MW & 12.56M & 8.37M & W1024MW-a & W1024MW-b \\
\bottomrule
\end{tabular}

\end{table}

\noindent\textbf{Non-grid benchmarks.}
To assess \acronym's generalisation beyond spatial models, we evaluated two different MDP families.
The first type of model is a \emph{sequential arena} with several configurations of ($K\in\{8,16\}$ arenas,
$\sim\!5\!\times\!10^{4}$ to $\sim\!8\!\times\!10^{5}$ states). These
represent multi-phase SSP problems in which an agent must complete $K$ arenas in sequence.
Moreover, each arena evolves independently 
(selected by a global counter), with local progress variables. The interaction between
the various arenas is limited and happens only at transition points.
Our goal is to minimise expected steps to completion: $\RR_{\min}[\,\mathsf{F}\,\texttt{finished}\,]$.

 In addition, we evaluate the well-known \emph{protocol} models such as the IEEE 802.11 wireless LAN (\texttt{wlan3}, $\sim\!97$K
 states) and the IEEE 1394 FireWire root contention protocol (\texttt{firewire}, $\sim\!4$K states). The first model involves three stations competing for medium
access via exponential backoff with shared channel variables.
The \texttt{firewire} model represents the tree identification phase where two nodes compete to become root through randomised leader election. 
For these models, we check $\PP_{\max}[\,\textit{true}\ \mathsf{U}\ (\texttt{bc1}{=}3 \lor \texttt{bc2}{=}3)\,]$
and minimise expected time to leader election for \texttt{wlan3} and \texttt{firewire}, respectively.
Additionally, we evaluate the \emph{consensus} shared coin protocol ($\sim\!22$K states, $\PP_{\max}$) and the \emph{zeroconf} network configuration protocol ($\sim\!1$K states, $\RR_{\min}$). All four protocol models are from the PRISM benchmark suite~\cite{prism}.

\textbf{Baselines and configurations.}
We compare \acronym\ against PRISM's explicit engine using VI and GS with/without pre-computations~\cite{prism}.
As \acronym\ extends PRISM in Java, comparing against PRISM's engine isolates algorithmic improvements from implementation differences (unlike comparing to the C++-based Storm~\cite{storm} probabilistic model checker).

We explored \acronym\ configurations with: maximum depth $D \in \{1, 2, 3, 4, 5\}$; initial partition sizes from $8 \times 8$ to $25 \times 25$; cost thresholds $\theta \in \{200, 300, 400, 500, 600, 800\}$;
and probability thresholds $\theta \in \{0.2, 0.3, 0.5, 0.7, 1.0\}$.
Unless stated otherwise, all sub-MDPs were solved using VI (convergence threshold $\varepsilon=10^{-6}$).
For depth-adapted refinement thresholds, we used $\theta_d = \theta$ for probability objectives and $\theta_d = \max(0.01, \theta/(10(1+d)))$ for cost objectives.
The boundary-change threshold was set to $\eta_{\mathrm{thr}} = 0.001$ for probability objectives and $\eta_{\mathrm{thr}} = 10^{-6}$ for cost objectives.
High uncertainty blocks (for probability objectives) were identified as those with average values in $(0.1, 0.9)$.
The numerical stability parameter in the refinement score formula was set to $\beta = 10^{-6}$.

\noindent\textbf{Objective-specific refinement score.}
In our experiments, we used different refinement scores depending on the verification objective.
For the warehouse (grid) experiments, we used the absolute score $\tau=\Delta$, together with
$\theta \in \{200,300,400,500,600,800\}$, measured in expected steps.
For the non-grid benchmarks, we used the normalised score
$\tau=\Delta/(|\mathrm{avg}(V)|+\beta)$, together with small threshold values (e.g., $0.05$).
In the results, we report, for each instance, the configuration that achieved the best execution time,
and, where relevant, we indicate the corresponding value of $\theta$ in the captions.

\begin{figure}[!t]
    \centering
    \begin{subfigure}[b]{0.48\textwidth}
        \centering
        \includegraphics[width=\textwidth]{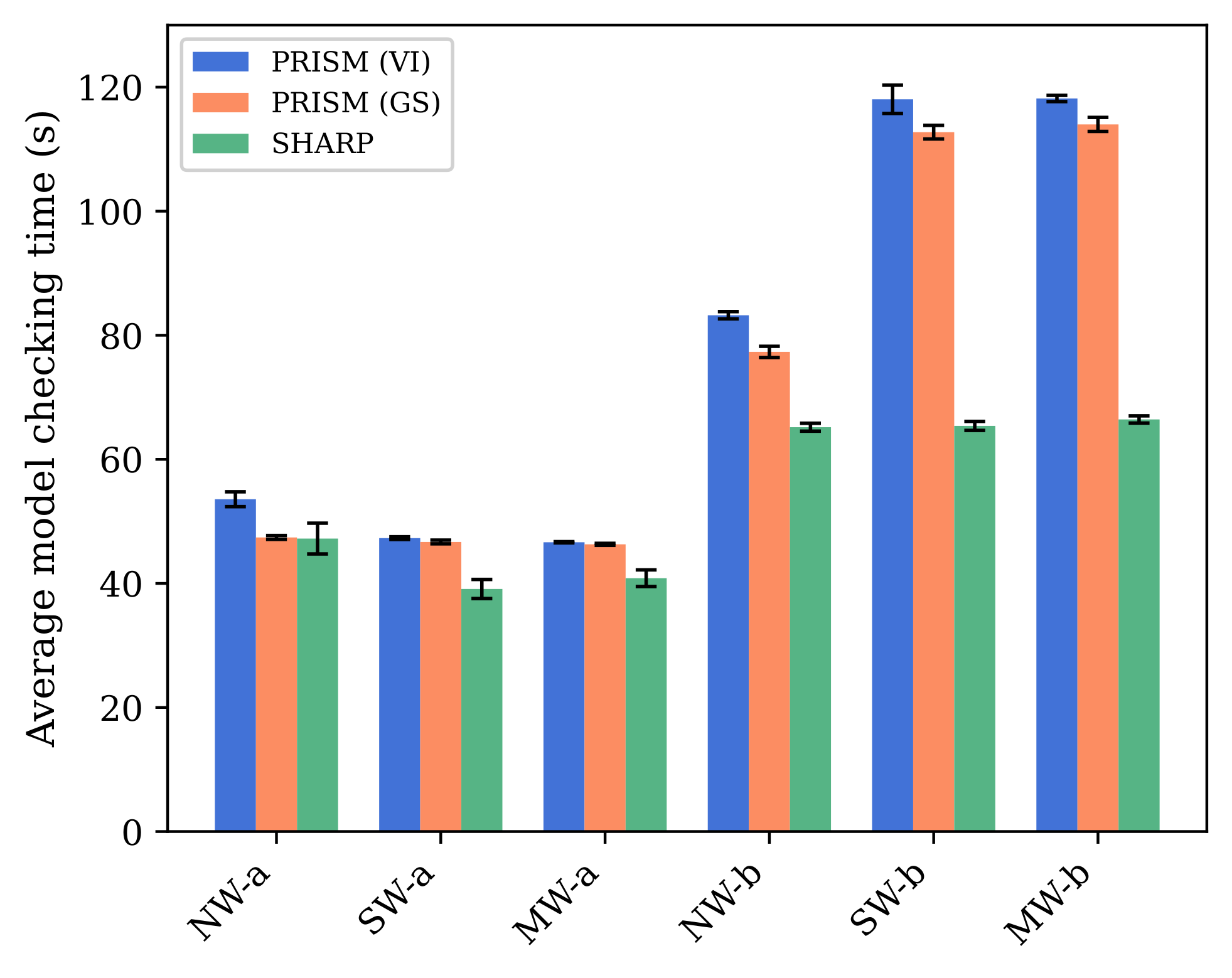}
    \end{subfigure}
    \hfill
    \begin{subfigure}[b]{0.48\textwidth}
        \centering
        \includegraphics[width=\textwidth]{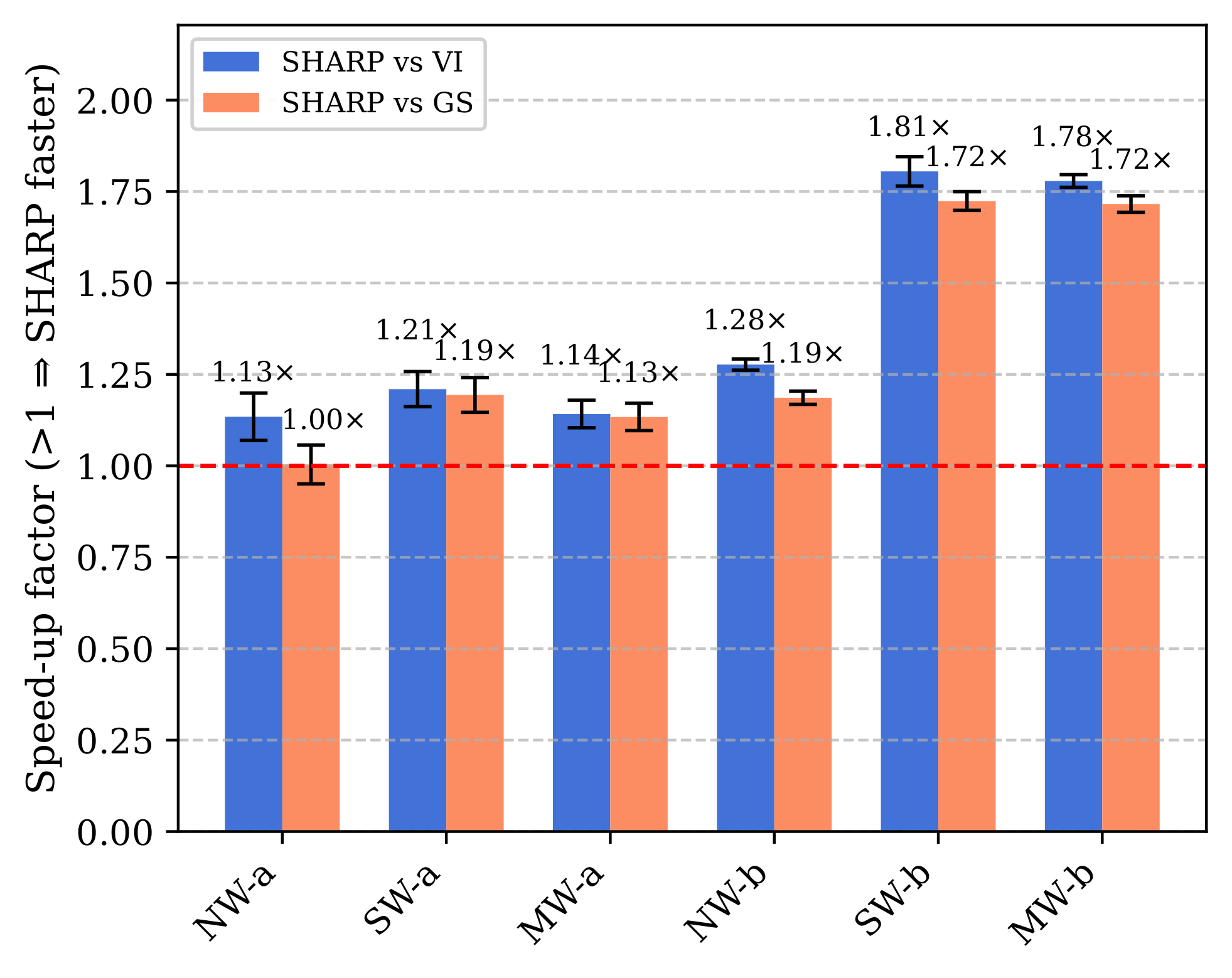}
    \end{subfigure}
    \caption{\acronym\ vs.\ PRISM on $1024\times1024$ warehouse instances: model-checking time (left) and speedup (right).}
    \Description{Two side-by-side charts comparing SHARP with PRISM on 1024 by 1024 warehouse benchmarks. The left chart shows absolute model-checking times for several instances. The right chart shows SHARP speedup over PRISM VI and GS on the same instances.}
    \label{fig:performance-comparison}
\end{figure}

\textbf{Experimental environment.}
We conducted the experiments on a laptop with \qty{64}{\giga\byte} of RAM and a \qty{5}{\giga\hertz} Intel 13th Gen i7-13700H processor running Ubuntu 22.04.5 LTS (64-bit). 
To develop and evaluate \acronym, we used OpenJDK 23.0.2 and PRISM 4.8.1~\cite{prism}. 
In all experiments, we set a timeout of \qty{30}{\minute} and a memory limit of \qty{32}{\giga\byte}.
Also, to mitigate any potential effect of randomisation, we run all experiments 30 times
and report mean$\pm$std where applicable.

\subsection{Results \& Discussion}

\noindent
\textbf{RQ1 (Efficiency).} 
We assess \acronym's efficiency against PRISM~\cite{prism} by measuring the model checking time and memory overhead across the case studies from Table~\ref{tab:case_studies}. In Table~\ref{tab:results}, we compare SHARP with PRISM's explicit-engine methods, VI and Gauss-Seidel (GS), both with and without precomputation on the two properties described earlier.

We observe that across the smaller model instances (e.g., W512NW-a/b), PRISM yields a policy in approximately \qtyrange{6}{15}{\second}, using \qtyrange{1.6}{1.9}{\giga\byte} of memory. On the other hand, SHARP yields the policy in about \qtyrange{8}{13}{\second} and uses \qtyrange{6.4}{8.5}{\giga\byte}. For the W512SW-a/b and W512MW-a/b instances, both PRISM and \acronym\ finish in less than \qty{20}{\second}, although PRISM is sometimes faster when no precomputation is used.
However, in larger problem instances with over 1 million states, such as the W1024* ones, PRISM-VI typically takes \qtyrange{46}{118}{\second} and uses at least \qty{5}{\giga\byte} of memory, whereas PRISM-GS does not consistently perform better, except in a few scenarios. Instead, \acronym's hierarchical analysis completes the $\PP_{\max}$ instances in approximately \qtyrange{39}{48}{\second} and the $\RR_{\min}$ instances in approximately \qtyrange{65}{67}{\second} (Figure~\ref{fig:performance-comparison} and Table~\ref{tab:results}).

For example, in \(\mathrm{W1024SW\text{-}a}\) ($\PP_{\max}$), PRISM-VI requires around \qty{47.3}{\second}, whereas \acronym\ takes only about \qty{39.1}{\second}, resulting in a saving of over \qty{8}{\second}. For \(\mathrm{W1024SW\text{-}b}\) ($\RR_{\min}$), PRISM-VI/GS need approximately \qty{118}{\second}/\qty{113}{\second}, while \acronym\ requires \qty{65.4}{\second}, achieving a 1.7\(\times\) speedup.

\begin{table*}[!t]
\centering
\caption{Comparison of \acronym\
against PRISM using Value Iteration (VI) and Gauss-Seidel (GS) for policy synthesis.\
\emph{Note:} $\theta$ is the configured refinement threshold (Pmax: probability difference; Rmin: absolute spread in expected steps).}
\label{tab:results}
\resizebox{\textwidth}{!}{
\begin{tabular}{l | rr r rr r | rr r rlr}
\toprule
& \multicolumn{6}{c|}{\textbf{PRISM}} & \multicolumn{6}{c}{\textbf{SHARP}} \\
\cmidrule(lr){2-7} \cmidrule(lr){8-13}
& \multicolumn{3}{c}{\textbf{VI}} & \multicolumn{3}{c|}{\textbf{GS}} & \multicolumn{6}{c}{} \\
\cmidrule(lr){2-4} \cmidrule(lr){5-7}
\textbf{ID} & \textbf{T (s) (std)} & \textbf{M (MBs)} & \textbf{R} & \textbf{T (s) (std)} & \textbf{M (MBs)} & \textbf{R} & \textbf{T (s) (std)} & \textbf{M (MBs)} & \textbf{R} & \textbf{D} & \textbf{$\theta$} &$\!\!\!\!$\shortstack{\textbf{Blocks}\\\textbf{X/Y}} \\
\midrule
\multicolumn{13}{c}{\textbf{Property: Pmax = [F ``goal'']}} \\
\midrule
$\!\!\!$W512NW-a (nopre) 
  & $6.27~(\pm0.27)$ & $1901~(\pm29)$ & 0.567 
  & $6.14~(\pm0.13)$ & $1899~(\pm40)$ & 0.567 
  &\multirow{2}{*}{$7.99~(\pm0.37)$} 
  & \multirow{2}{*}{$8371~(\pm207)$} 
  & \multirow{2}{*}{$0.567$} 
  & \multirow{2}{*}{$1$} 
  & \multirow{2}{*}{$0.5$} 
  & \multirow{2}{*}{$8/8$} 
\\
$\!\!\!$W512NW-a (pre)$\!\!$ 
  & $15.69~(\pm0.32)$ & $1907~(\pm27)$ & 0.567 
  & $15.54~(\pm0.24)$ & $1905~(\pm18)$ & 0.567 
  &
  &
  &
  &
  &
  &
\\

$\!\!\!$W512SW-a (nopre)$\!\!$ 
  & $5.96~(\pm0.04)$ & $1889~(\pm19)$ & 0.567
  & $5.94~(\pm0.05)$ & $1889~(\pm14)$ & 0.567
  & \multirow{2}{*}{$8.29~(\pm0.43)$}
  & \multirow{2}{*}{$8369~(\pm438)$}
  & \multirow{2}{*}{$0.567$}
  & \multirow{2}{*}{$1$}
  & \multirow{2}{*}{$0.5$}
  & \multirow{2}{*}{$8/8$}
\\
$\!\!\!$W512SW-a (pre)$\!\!$ 
  & $15.72~(\pm0.13)$ & $1912~(\pm24)$ & 0.567
  & $15.27~(\pm0.11)$ & $1900~(\pm15)$ & 0.567
  &
  &
  &
  &
  &
  &
\\

$\!\!\!$W512MW-a (nopre)$\!\!$ 
  & $6.01~(\pm0.05)$ & $1928~(\pm26)$ & 0.567 
  & $6.05~(\pm0.04)$ & $1926~(\pm55)$ & 0.567
  & \multirow{2}{*}{$7.61~(\pm0.31)$}
  & \multirow{2}{*}{$8506~(\pm264)$}
  & \multirow{2}{*}{$0.567$}
  & \multirow{2}{*}{$1$}
  & \multirow{2}{*}{$0.5$}
  & \multirow{2}{*}{$8/8$}
\\
$\!\!\!$W512MW-a (pre)$\!\!$ 
  & $15.75~(\pm0.19)$ & $1935~(\pm57)$ & 0.567 
  & $15.62~(\pm0.22)$ & $1932~(\pm53)$ & 0.567
  &
  &
  &
  &
  &
  &
\\

\midrule
$\!\!\!$W1024NW-a (nopre)$\!\!$
  & $53.57~(\pm1.20)$ & $6213~(\pm98)$ & 0.278
  & $47.41~(\pm0.31)$ & $6204~(\pm156)$ & 0.278
  &$\!$\multirow{2}{*}{$47.23~(\pm2.48)$}
  & \multirow{2}{*}{$19769~(\pm1660)$}
  & \multirow{2}{*}{$0.278$}
  & \multirow{2}{*}{$1$}
  & \multirow{2}{*}{$0.5$}
  & \multirow{2}{*}{$8/8$}
\\
$\!\!\!$W1024NW-a (pre)$\!\!$
  & $129.71~(\pm1.92)$ & $6209~(\pm95)$ & 0.278
  & $123.41~(\pm3.21)$ & $6228~(\pm52)$ & 0.278
  &
  &
  &
  &
  &
  &
\\

$\!\!\!$W1024SW-a (nopre)$\!\!$
  & $47.30~(\pm0.22)$ & $6268~(\pm128)$ & 0.321
  & $46.68~(\pm0.30)$ & $6242~(\pm93)$ & 0.321
  &$\!$\multirow{2}{*}{$39.10~(\pm1.54)$}
  & \multirow{2}{*}{$20029~(\pm1598)$}
  & \multirow{2}{*}{$0.321$}
  & \multirow{2}{*}{$1$}
  & \multirow{2}{*}{$1.0$}
  & \multirow{2}{*}{$8/8$}
\\
$\!\!\!$W1024SW-a (pre)$\!\!$
  & $125.55~(\pm0.73)$ & $6262~(\pm87)$ & 0.321
  & $122.02~(\pm1.04)$ & $6300~(\pm56)$ & 0.321
  &
  &
  &
  &
  &
  &
\\

$\!\!\!$W1024MW-a (nopre)$\!\!$
  & $46.63~(\pm0.10)$ & $6612~(\pm58)$ & 0.321
  & $46.30~(\pm0.16)$ & $6597~(\pm96)$ & 0.321
  &$\!$\multirow{2}{*}{$40.84~(\pm1.34)$}
  & \multirow{2}{*}{$19339~(\pm729)$}
  & \multirow{2}{*}{$0.321$}
  & \multirow{2}{*}{$1$}
  & \multirow{2}{*}{$0.5$}
  & \multirow{2}{*}{$8/8$}
\\
$\!\!\!$W1024MW-a (pre)$\!\!$
  & $125.17~(\pm0.84)$ & $6608~(\pm104)$ & 0.321
  & $121.75~(\pm0.85)$ & $6630~(\pm45)$ & 0.321
  &
  &
  &
  &
  &
  &
\\

\midrule
\midrule
\multicolumn{13}{c}{\textbf{Property: R\{\text{``steps''}\}\text{min=? [F ``goal'']}}} \\
\midrule
$\!\!\!$W512NW-b (nopre)$\!\!$
  & $10.92~(\pm0.20)$ & $1624~(\pm48)$ & $1277.49$
  & $10.36~(\pm0.30)$ & $1615~(\pm41)$ & $1277.50$
  &$\!$\multirow{2}{*}{$12.58~(\pm0.19)$}
  & \multirow{2}{*}{$6385~(\pm163)$}
  & \multirow{2}{*}{$1277.50$}
  & \multirow{2}{*}{$1$}
  & \multirow{2}{*}{$500$}
  & \multirow{2}{*}{$8/8$}
\\
$\!\!\!$W512NW-b (pre)$\!\!$
  & $10.96~(\pm0.26)$ & $1609~(\pm37)$ & $1277.49$
  & $10.37~(\pm0.27)$ & $1604~(\pm38)$ & $1277.50$
  &
  &
  &
  &
  &
  &
\\

$\!\!\!$W512SW-b (nopre)$\!\!$
  & $14.93~(\pm0.15)$ & $1170~(\pm8)$ & $1277.49$
  & $14.28~(\pm0.14)$ & $1167~(\pm12)$ & $1277.50$
  &$\!$\multirow{2}{*}{$12.49~(\pm0.19)$}
  & \multirow{2}{*}{$6470~(\pm230)$}
  & \multirow{2}{*}{$1277.50$}
  & \multirow{2}{*}{$1$}
  & \multirow{2}{*}{$200$}
  & \multirow{2}{*}{$8/8$}
\\
$\!\!\!$W512SW-b (pre)$\!\!$
  & $14.93~(\pm0.17)$ & $1171~(\pm11)$ & $1277.49$
  & $14.26~(\pm0.13)$ & $1170~(\pm10)$ & $1277.50$
  &
  &
  &
  &
  &
  &
\\

$\!\!\!$W512MW-b (nopre)$\!\!$
  & $15.51~(\pm0.30)$ & $1741~(\pm23)$ & $1277.49$
  & $14.96~(\pm0.28)$ & $1735~(\pm28)$ & $1277.50$
  &$\!$\multirow{2}{*}{$12.65~(\pm0.19)$}
  & \multirow{2}{*}{$6453~(\pm293)$}
  & \multirow{2}{*}{$1277.50$}
  & \multirow{2}{*}{$1$}
  & \multirow{2}{*}{$200$}
  & \multirow{2}{*}{$8/8$}
\\
$\!\!\!$W512MW-b (pre)$\!\!$
  & $15.42~(\pm0.22)$ & $1734~(\pm35)$ & $1277.49$
  & $14.70~(\pm0.29)$ & $1744~(\pm31)$ & $1277.50$
  &
  &
  &
  &
  &
  &
\\

\midrule
$\!\!\!$W1024NW-b (nopre)$\!\!$
  & $83.23~(\pm0.58)$ & $5358~(\pm60)$ & $2557.48$
  & $77.32~(\pm0.90)$ & $5360~(\pm84)$ & $2557.50$
  &$\!$\multirow{2}{*}{$65.18~(\pm0.64)$}
  & \multirow{2}{*}{$18179~(\pm1048)$}
  & \multirow{2}{*}{$2557.50$}
  & \multirow{2}{*}{$1$}
  & \multirow{2}{*}{$200$}
  & \multirow{2}{*}{$8/8$}
\\
$\!\!\!$W1024NW-b (pre)$\!\!$
  & $83.29~(\pm0.36)$ & $5344~(\pm32)$ & $2557.48$
  & $77.80~(\pm1.25)$  & $5363~(\pm85)$ & $2557.50$
  &
  &
  &
  &
  &
  &
\\
$\!\!\!$W1024SW-b (nopre)$\!\!$
  & $118.04~(\pm2.28)$ & $5427~(\pm64)$ & 2557.48
  & $112.74~(\pm1.10)$ & $5412~(\pm33)$ & 2557.50
  &$\!\!$\multirow{2}{*}{$65.39~(\pm0.73)$}
  & \multirow{2}{*}{$18297~(\pm1027)$}
  & \multirow{2}{*}{$2557.50$}
  & \multirow{2}{*}{$1$}
  & \multirow{2}{*}{$500$}
  & \multirow{2}{*}{$8/8$}
\\
$\!\!\!$W1024SW-b (pre)$\!\!$
  & $117.83~(\pm1.03)$ & $5491~(\pm172)$ & 2557.48
  & $113.29~(\pm1.53)$ & $5410~(\pm29)$ & 2557.50
  &
  &
  &
  &
  &
  &
\\

$\!\!\!$W1024MW-b (nopre)$\!\!$
  & $118.17~(\pm0.50)$ & $5808~(\pm38)$ & $2557.48$
  & $113.99~(\pm1.13)$ & $5778~(\pm97)$ & $2557.50$
  &$\!$\multirow{2}{*}{$66.43~(\pm0.58)$}
  & \multirow{2}{*}{$16938~(\pm1337)$}
  & \multirow{2}{*}{$2557.50$}
  & \multirow{2}{*}{$1$}
  & \multirow{2}{*}{$500$}
  & \multirow{2}{*}{$8/8$}
\\
$\!\!\!$W1024MW-b (pre)$\!\!$
  & $118.52~(\pm1.05)$ & $5814~(\pm52)$ & $2557.48$
  & $114.37~(\pm2.00)$ & $5796~(\pm105)$ & $2557.50$
  &
  &
  &
  &
  &
  &
\\
\bottomrule
\end{tabular}
}
\end{table*}

Note that even in the $\PP_{\max}$ cases where the performance gap is smaller than the \qty{8}{\second} difference observed in \(\mathrm{W1024SW\text{-}a}\), SHARP still offers a noticeable speedup of about 1.2$\times$. 
For instance, in \(\mathrm{W1024MW\text{-}a}\), PRISM takes $\approx$46.3--46.6 s, while SHARP completes in $\approx$40.8 s.

We note that although \acronym\ may require 17--20\,GB of memory on our million-state instances (Table~\ref{tab:results}),
this remains within typical machine limits.
The footprint, it turns out, has two components.
Some of the extra memory is inherent to \acronym, since we keep the partition
tree and per-leaf data. In our implementation, this part is about 12\,bytes per
state. Most of the memory we observe, however, comes from PRISM's Java-side
representation, especially object headers and the maps used by the collections.

\begin{figure*}[t]
    \centering
    \includegraphics[width=\textwidth]{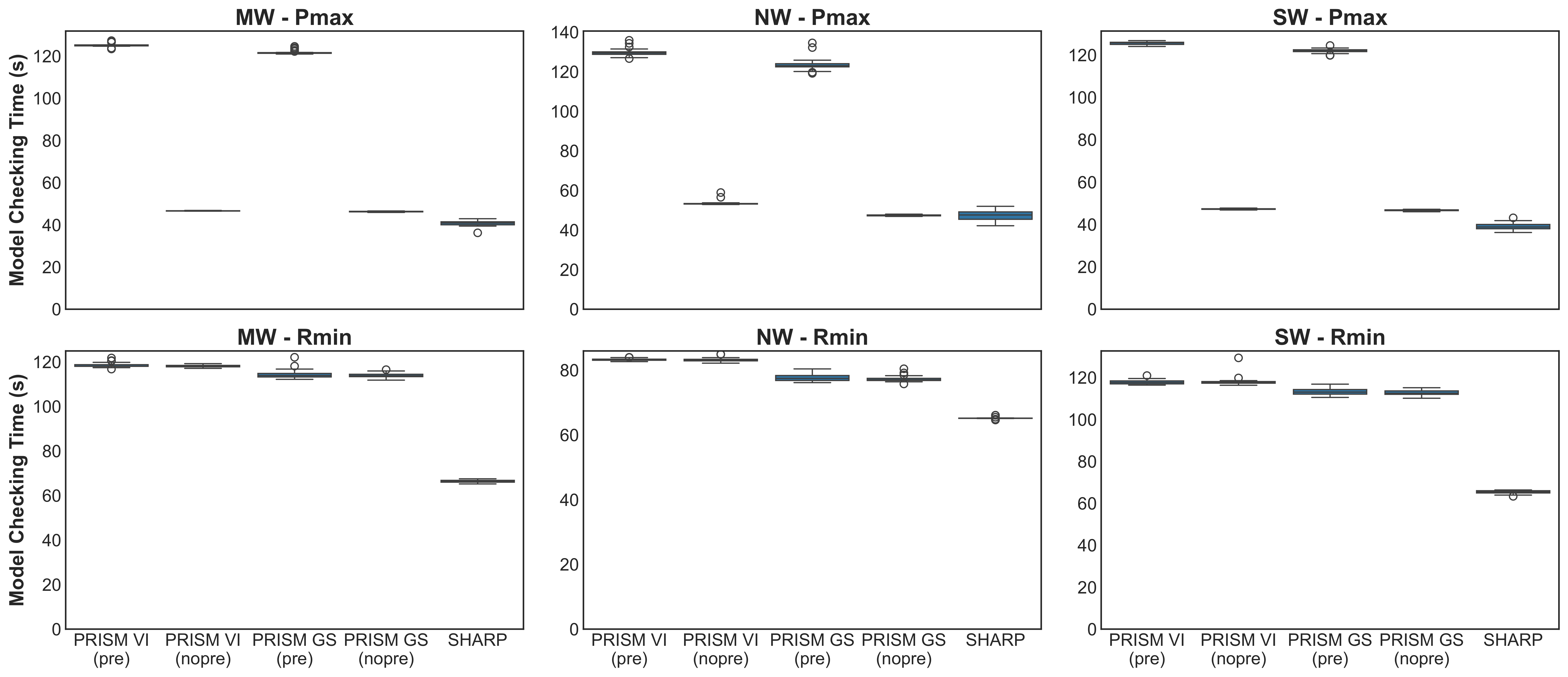}
    \caption{Execution time comparison: \acronym\ vs.\ PRISM (VI/GS) on W1024* models for Pmax and Rmin properties over 30 independent runs.}
    \Description{A grouped set of boxplots comparing SHARP, PRISM VI, and PRISM GS on the 1024 by 1024 warehouse benchmarks. The plot shows execution-time distributions over 30 runs for six benchmark and property combinations.}
    \label{fig:boxplot-prism-sharp-detailed}
\end{figure*}

Also, differences in solution quality from PRISM are negligible; the largest gap we measure is $1.3\times 10^{-4}$ (see Section~\ref{sec:evaluation}, RQ4 for a detailed analysis).
\acronym's median execution time is lower than PRISM's in 5 out of 6 cases, with the largest gains on $\RR_{\min}$ (Figure~\ref{fig:boxplot-prism-sharp-detailed}).
On our largest warehouse instances, we observe about $1.2$--$2\!\times$ speedup, with greater improvements for reward objectives than probability objectives. For time-sensitive tasks, these gains can outweigh the increased memory overhead.

\begin{wrapfigure}[14]{R}{0.33\columnwidth}
    \centering
    \includegraphics[width=0.30\columnwidth]{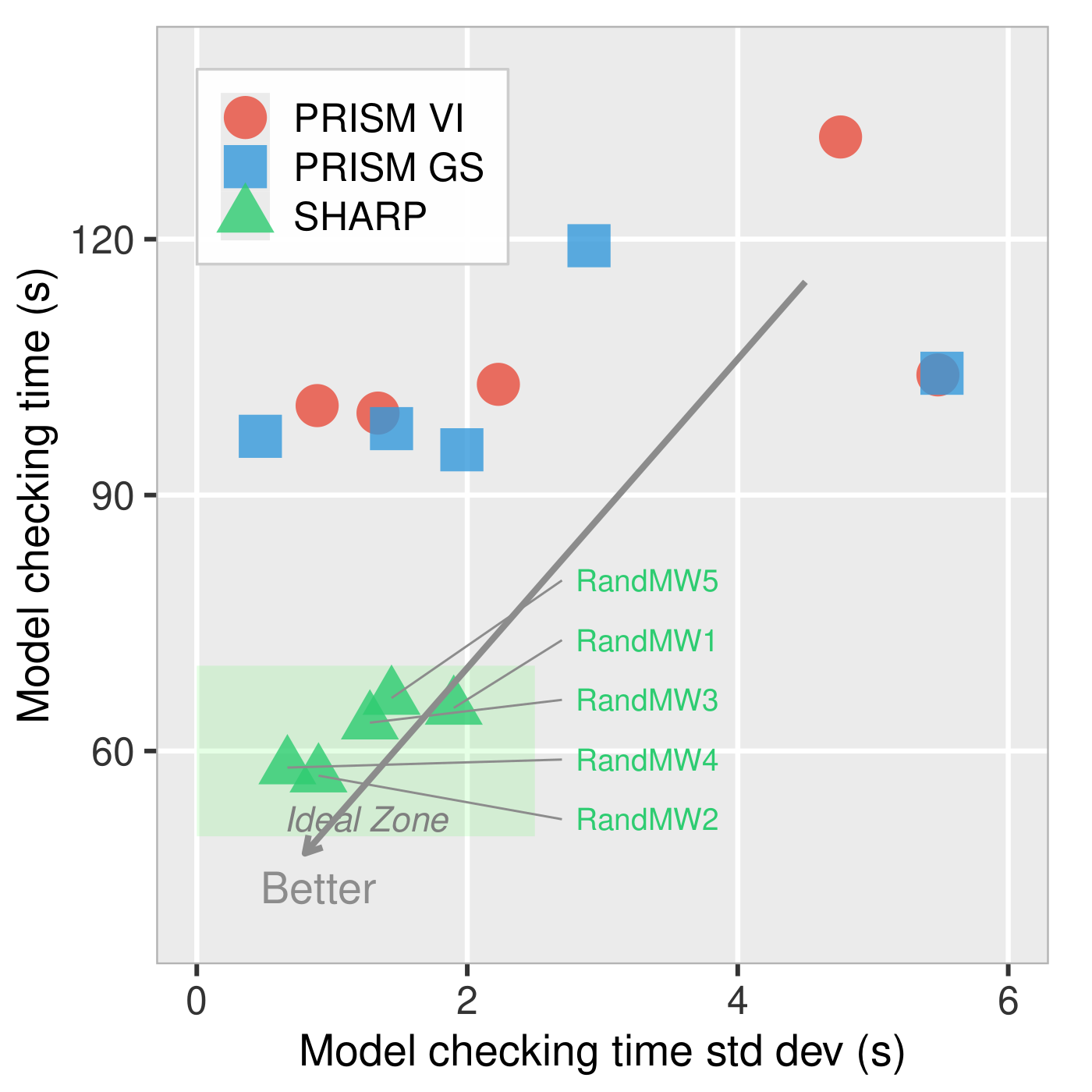}
    \caption{Performance stability on random $1024\times1024$ MW models.}
    \Description{A scatter plot comparing SHARP, PRISM VI, and PRISM GS on five random 1024 by 1024 warehouse models. The vertical axis is mean runtime and the horizontal axis is runtime standard deviation, highlighting that SHARP is both faster and more stable.}
    \label{fig:random-performance}
\end{wrapfigure}

\textbf{Randomised robustness experiments.}
Figure~\ref{fig:random-performance}
assesses all methods on five 1024$\times$1024 models whose wall segments were placed at random positions (RandMW1--RandMW5). 
The points show the mean verification time (y-axis) against the run-to-run standard deviation (x-axis). 
The lower-left corner represents solutions that are both fast and stable. 
Across all five randomised problem instances, \acronym\ forms a tight cluster inside the highlighted ``ideal block'', with means between \qtyrange{57}{66}{\second} and standard deviation below \qty{2}{\second}. 
PRISM-GS is consistently slower and more variable, while PRISM-VI is slower still, often taking nearly twice as long as \acronym\ and exhibiting up to three times its standard deviation. 
The clear separation confirms that \acronym\ delivers not only faster verification on large grid MDPs with randomly-placed walls, but also far more predictable performance across different random layouts.

\begin{figure}[t]
    \centering
    \begin{subfigure}[b]{0.49\columnwidth}
        \centering
        \includegraphics[width=\textwidth]{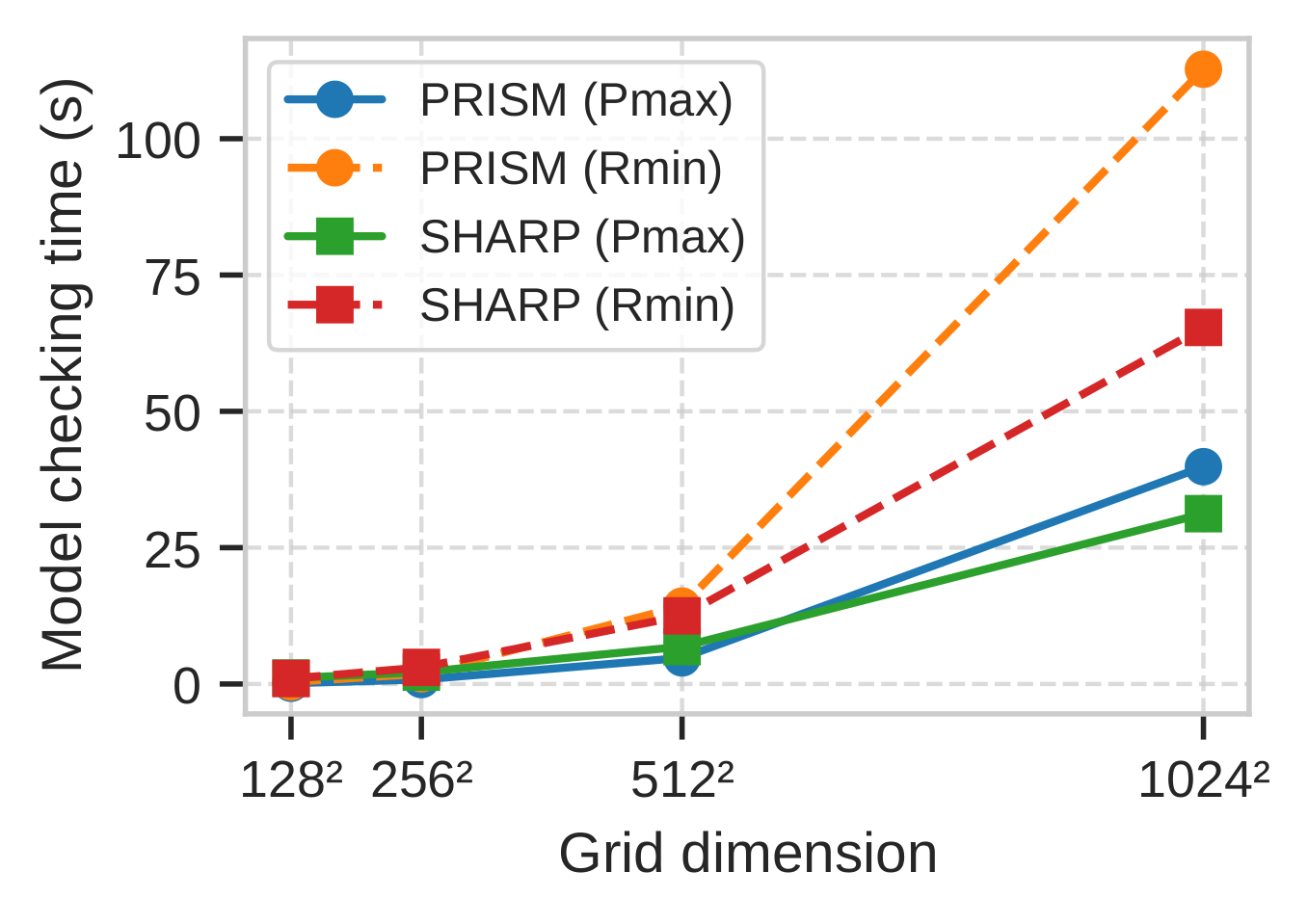}
    \end{subfigure}
    \hfill
    \begin{subfigure}[b]{0.49\columnwidth}
        \centering
        \includegraphics[width=\textwidth]{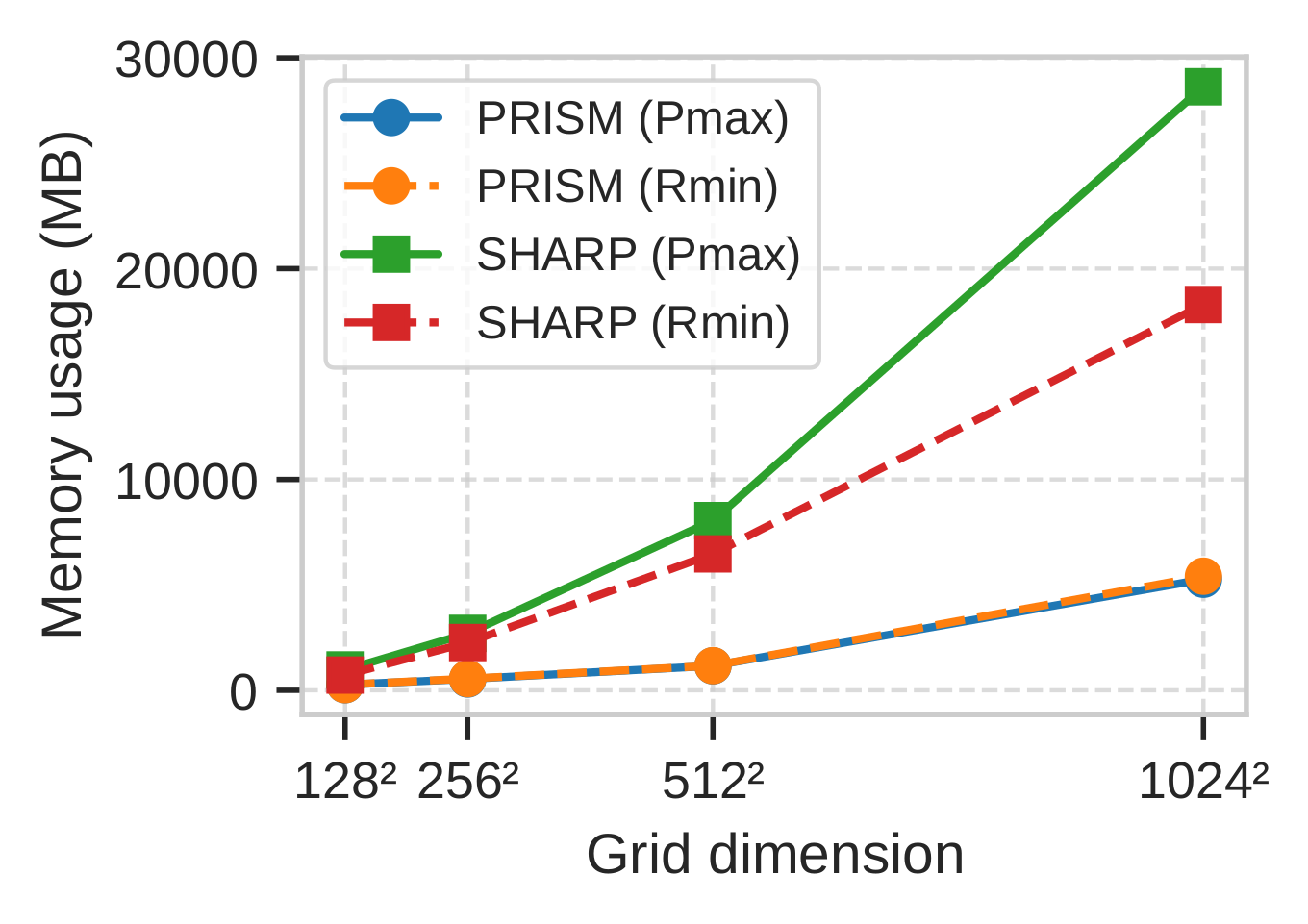}
    \end{subfigure}
    \caption{Scalability comparison: time (left) and memory (right) for PRISM vs.\ \acronym\ across grid dimensions.}
    \Description{Two line charts comparing PRISM and SHARP across increasing grid sizes. The left chart shows model-checking time, and the right chart shows memory usage, highlighting SHARP's better runtime scaling but higher memory footprint.}
    \label{fig:scalability}
\end{figure}

\textbf{RQ2 (Scalability).}
  We evaluate scalability by comparing \acronym\ against the best-performing PRISM configuration (GS nopre) on the SW warehouse model
  across grid sizes $128\!\times\!128$ to $1024\!\times\!1024$ (Figure~\ref{fig:scalability}).                                           
  For smaller grids (up to $256\times256$), \acronym's overhead outweighs its benefits.
  However, at $512\times512$ grids ($2.6\times10^5$ states), \acronym\ already performs better on $\RR_{\min}$ (12.49s vs 14.28s), and by
   $1024\times1024$ ($10^6$ states) it is faster on both properties: $\PP_{\max}$ in 39.10s vs 46.68s and $\RR_{\min}$ in 65.39s vs
  112.74s.
  Since \acronym\ maintains additional data structures for hierarchical refinement, its memory footprint is unsurprisingly
  higher across all grid sizes (Figure~\ref{fig:scalability}, right); for instance, 8.2\,GB vs 1.1--1.9\,GB at $512\times512$.
  Considering the memory capabilities of modern machines, these overheads are acceptable and well within reasonable limits.
  Across our grid-scaling experiment (Figure~\ref{fig:scalability}), \acronym's runtime grows substantially more slowly with model size than PRISM's; e.g., from $512^2$ to $1024^2$ ($4\times$ states), \acronym\ increases by about $5\times$ while PRISM increases by about $8\times$, meaning that
  \acronym's scaling advantage is valuable for engineers verifying growing models.
  As we discuss in the guidance section, this scaling advantage is most pronounced on models with spatial or staged-progression
  structure.

\textbf{RQ3 (Generalisation).}
We used diverse MDP structures to evaluate \acronym's generalisation beyond spatial models.
These include sequential arena models, the wlan3 network protocol, the firewire root contention protocol, the consensus shared coin protocol, and the zeroconf network configuration protocol from the PRISM benchmark suite, using two different partition strategies.
For \texttt{arenas8}, we used SCC-based partitioning; for \texttt{arenas16}, we grouped states by their progress counter, putting consecutive values into the same group (e.g., 0-3, 4-7, etc.). This enabled solving each group as a sub-MDP, only refining groups when needed.

Furthermore, for protocol models, we used SCC-based partitioning for \texttt{wlan3}, \texttt{consensus}, and \texttt{zeroconf}, and counter-based partitioning for \texttt{firewire}.
SCC-based partitioning groups strongly connected components (clusters of states that can reach each other through transitions) into blocks,
while counter-based partitioning groups states by protocol progress variables.

In Table~\ref{tab:generalization}, we observe that, on the large sequential arena model (\texttt{arenas16} with 788K states), \acronym\ achieves nearly $2\times$ speedup over PRISM's standard VI, reducing model checking time 
  from \qty{295}{\second} to \qty{153}{\second}. This occurs because there is an alignment between the arena's sequential structure and our hierarchical decomposition approach, where each stage operates independently with minimal inter-block communication.
  
 On the other hand, \acronym\ did not perform well on protocol models, something which is evident
  from the model checking time which was $100\times$ and $5\times$ slower compared to PRISM, on \texttt{wlan3} and
  on \texttt{firewire}, respectively, and $5.6\times$--$7.4\times$ slower on \texttt{consensus} and \texttt{zeroconf}.
  This occurs because these protocol models have many inter-block dependencies due to the shared state variables and the
  requirement for synchronisation.
   Hence, the overhead of maintaining the hierarchy and coordinating between blocks outweighs any decomposition benefits, making running standard VI on the flat MDP a superior approach.

 These findings demonstrate that \acronym\ generalises to diverse types of MDP models. Further, the results show that model structure is as critical as size for hierarchical approaches.
 
 \begin{table}[t]
  \caption{\acronym\ Performance Across Different MDP Structures. All experiments use $D=3$ and $\theta=0.05$.}
  \label{tab:generalization}
  \centering
  \small
  \begin{tabular}{lrrrrrl}
  \toprule
  Model & States & Type & PRISM & \acronym & Speedup & Partition \\
  \midrule
  arenas8 ($\RR_{\min}$)  & 53K   & Sequential  & 2.29s   & 2.94s   & $0.78\times$ & SCC \\
  wlan3 ($\PP_{\max}$)    & 97K   & Protocol    & 0.047s  & 5.28s   & $0.01\times$ & SCC \\
  firewire ($\RR_{\min}$) & 4K    & Protocol    & 0.038s  & 0.187s  & $0.20\times$ & Counter \\
  arenas16 ($\RR_{\min}$) & 788K  & Sequential  & 295.53s & 152.75s & $1.93\times$ & Counter \\
  consensus ($\PP_{\max}$) & 22K & Protocol & 0.735s & 5.414s & $0.14\times$ & SCC \\
  zeroconf ($\RR_{\min}$) & 1.1K & Protocol & 0.040s & 0.224s & $0.18\times$ & SCC \\
  \bottomrule
  \end{tabular}
  \end{table}

\begin{figure*}[t]
    \centering
    \begin{minipage}[t]{0.49\linewidth}
        \centering
        \includegraphics[width=\linewidth]{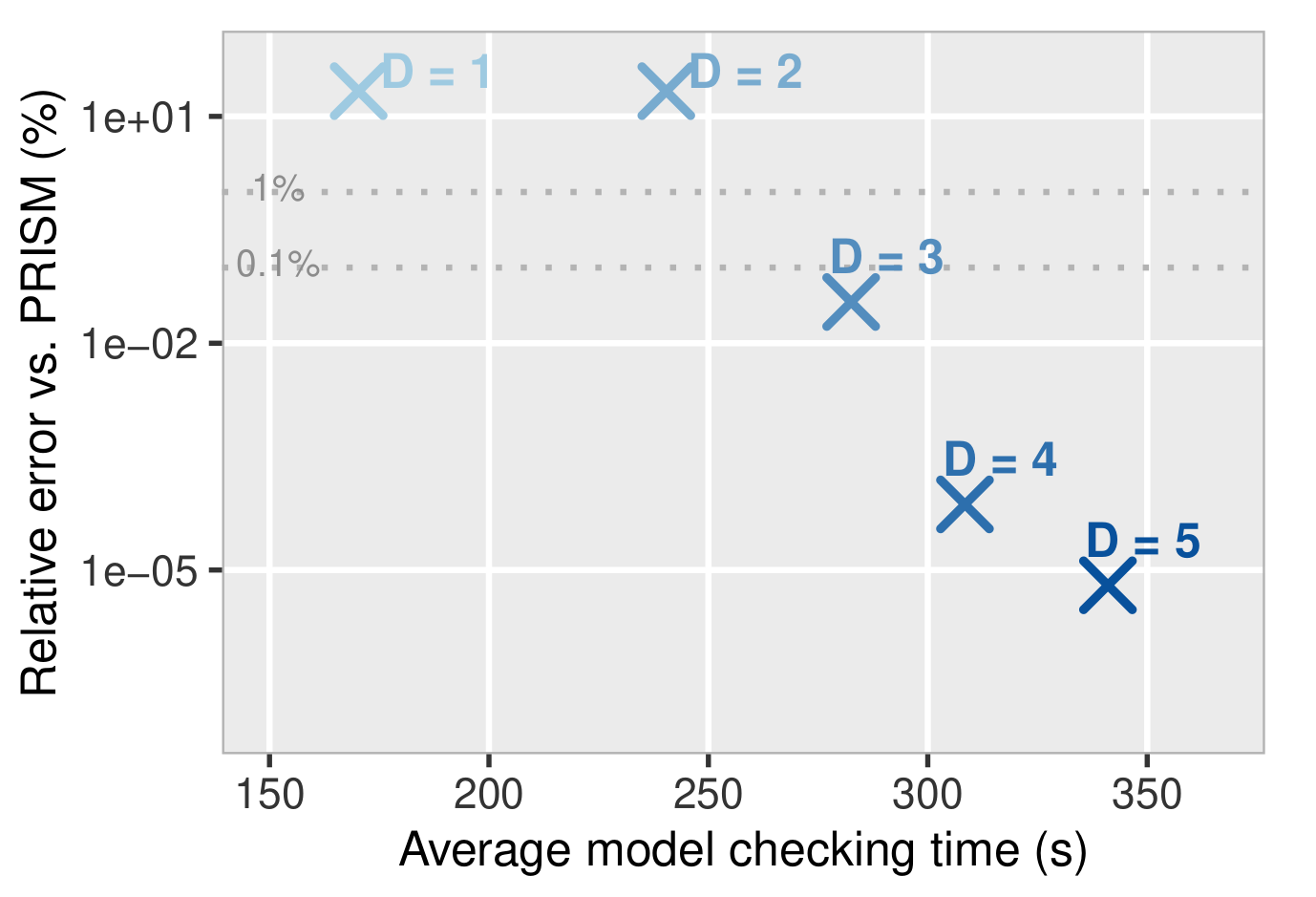}
        \caption{Policy quality vs.\ model-checking time across refinement depths ($D$) under capped local VI on 1024$\times$1024 MW ($\PP_{\max}$, $3{\times}3$ partition).}
        \label{fig:refinement-tradeoff}
    \end{minipage}
    \hfill
    \begin{minipage}[t]{0.49\linewidth}
        \centering
        \includegraphics[width=\linewidth]{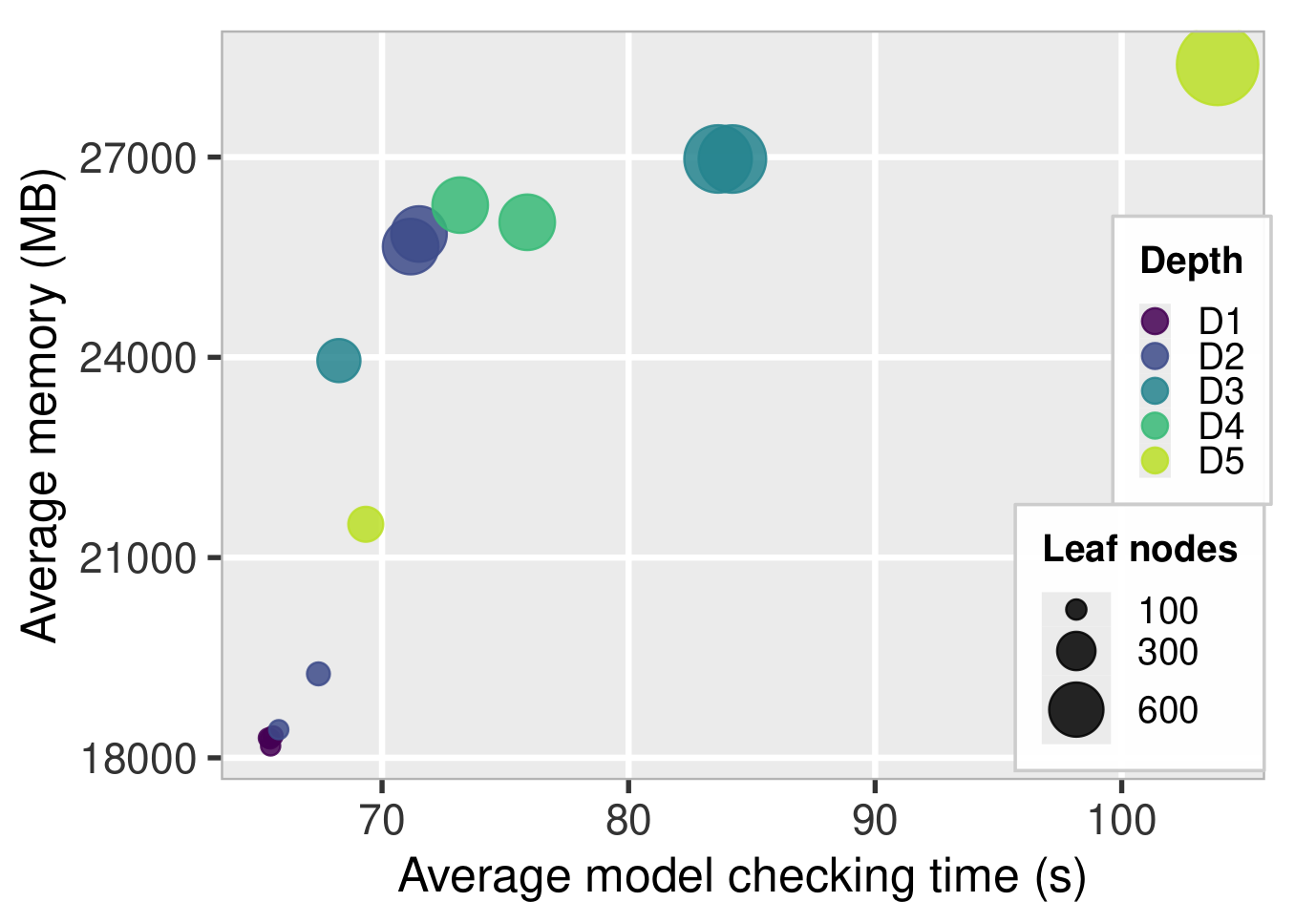}
        \caption{Scatter plot of SHARP hyperparameter configurations
          (Table~\ref{tab:config-params}) on the \(1024\times1024\) SW instance
          ($\RR_{\min}$).
          }
        \label{fig:hyper-param1}
    \end{minipage}
    \par\smallskip
    \centering
    \includegraphics[width=0.96\textwidth]{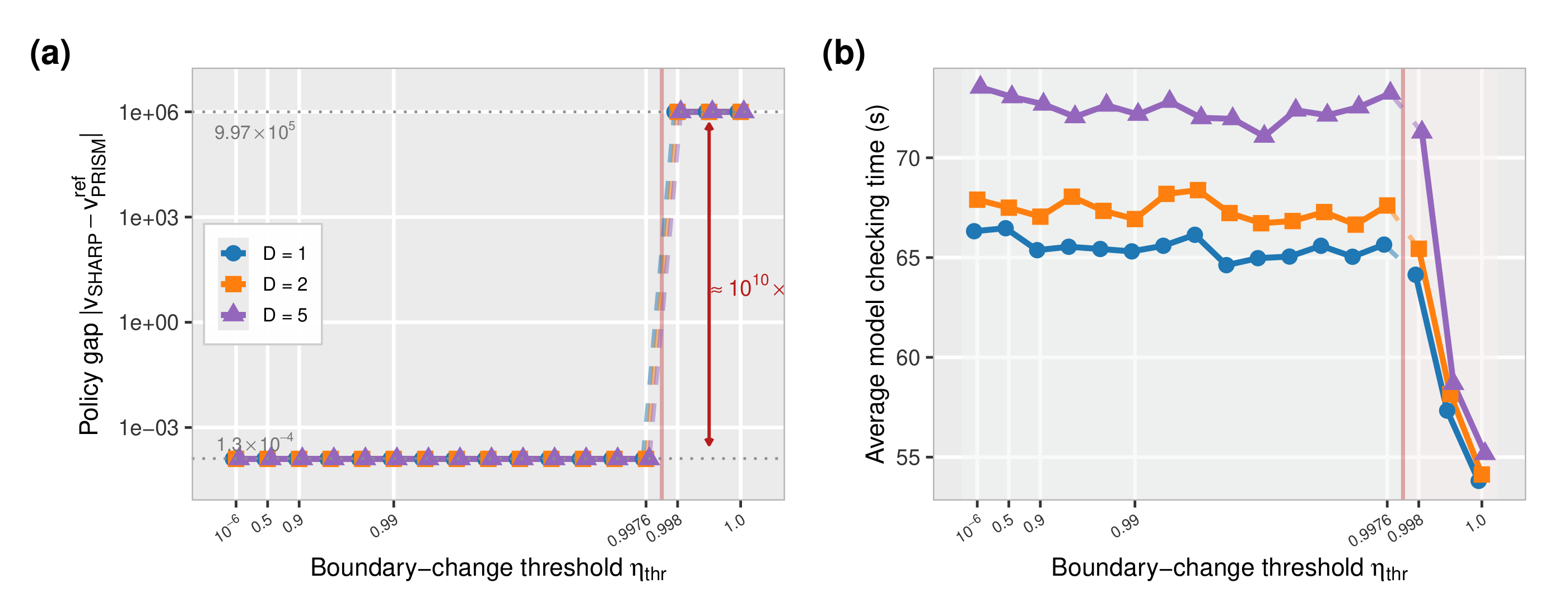}
    \caption{Effect of $\eta_{\mathrm{thr}}$ on 1024$\times$1024 SW ($\RR_{\min}$, $\theta{=}200$, $8{\times}8$): (a)~policy-quality gap and (b)~model-checking time.}
    \label{fig:eta-ablation}
    \Description{Three plots grouped at the top of one page. The top-left plot shows policy-quality gap versus model-checking time for different refinement depths under capped local VI. The top-right plot shows runtime and memory trade-offs for SHARP hyperparameter configurations on the 1024 by 1024 SW benchmark. The bottom plot has two coordinated panels showing the effect of the boundary-change threshold on the 1024 by 1024 SW benchmark, one for policy-quality gap and one for model-checking time.}
\end{figure*}

\textbf{RQ4 (Hyperparameter sensitivity).} 
We evaluate \acronym's sensitivity to key hyperparameters (maximum refinement depth $D$, cost threshold $\theta$,
and initial partition sizes) using the $1024\times1024$ SW instance ($\RR_{\min}$).
In Table~\ref{tab:config-params}, we show the performance impact for various configurations (A--E).
Moreover, in Figure~\ref{fig:hyper-param1} we plot each configuration's time and memory usage, and
colour the points according to the refinement depth $D$, where darker points indicate a lower $D$.
Results show that initial partition size dominates performance, more than refinement depth or threshold. The $D=1$ configurations (group A, $8\times8$ partitions) achieve the best performance with no refinements needed. While deeper refinements ($D$=2--5) increase resource usage, the initial partition size remains the critical factor: B1--B2 ($8\times8$) outperform B3--B4 ($16\times16$) even though B1 performs more refinements (5 vs 1); C2 ($20\times20$) requires 23\% more time than C1 ($12\times12$) with identical refinements; and E2 ($25\times25$) shows the worst overall performance. This pattern confirms that smaller initial partitions ($8\times8$ to $12\times12$) yield better performance regardless of refinement depth, as the overhead from managing numerous partitions outweighs refinement benefits.

\begin{table}[t]
\centering
\footnotesize
\caption{SHARP configuration parameters and performance on 1024$\times$1024 SW ($\RR_{\min}$).}
\label{tab:config-params}
\begin{tabular}{@{}lrrrr@{}}
\toprule
\textbf{Config} & \textbf{Time (s)} & \textbf{Mem (MB)} & \textbf{Leaves} & \textbf{Refinements} \\
\midrule
A1 (D1, 200, 8$\times$8)  & 65.6$\pm$0.8 & 18329$\pm$947  & 64  & 0 \\
A2 (D1, 500, 8$\times$8)  & 65.4$\pm$0.7 & 18297$\pm$1027 & 64  & 0 \\
A3 (D1, 300, 8$\times$8)  & 65.5$\pm$0.6 & 18283$\pm$1130 & 64  & 0 \\
A4 (D1, 600, 8$\times$8)  & 65.5$\pm$0.6 & 18178$\pm$962  & 64  & 0 \\
\midrule
B1 (D2, 200, 8$\times$8)  & 67.4$\pm$0.5 & 19258$\pm$1101 & 66  & 5 \\
B2 (D2, 500, 8$\times$8)  & 65.8$\pm$0.7 & 18422$\pm$980  & 64  & 1 \\
B3 (D2, 400, 16$\times$16) & 71.5$\pm$1.0 & 25848$\pm$523  & 256 & 1 \\
B4 (D2, 800, 16$\times$16) & 71.2$\pm$1.0 & 25659$\pm$685  & 256 & 1 \\
\midrule
C1 (D3, 300, 12$\times$12) & 68.3$\pm$0.6 & 23951$\pm$1259 & 147 & 5 \\
C2 (D3, 600, 20$\times$20) & 83.6$\pm$1.4 & 26972$\pm$562  & 403 & 5 \\
\midrule
D1 (D4, 400, 16$\times$16) & 73.2$\pm$1.4 & 26280$\pm$580  & 256 & 5 \\
D2 (D4, 600, 16$\times$16) & 75.9$\pm$1.4 & 26023$\pm$610  & 256 & 5 \\
\midrule
E1 (D5, 500, 10$\times$10) & 69.3$\pm$1.0 & 21499$\pm$1284 & 100 & 7 \\
E2 (D5, 800, 25$\times$25) & 103.9$\pm$3.7 & 28387$\pm$395 & 625 & 7 \\
\bottomrule
\end{tabular}
\end{table}

\paragraph{Policy Quality.}
In order to evaluate the quality of the policies produced by \acronym,
we measure the absolute gap $|V_{\acronym} - V^{\mathrm{ref}}_{\textsc{PRISM}}|$
between the value computed by \acronym\ and the reference value obtained from PRISM.
All 14~configurations listed in Table~\ref{tab:config-params} achieve the same gap
of $1.3\times 10^{-4}$. This is an important observation, as it confirms that
for the tested configurations and instances, the partitioning and refinement hyperparameters affect only time and memory,
not the quality of the computed policy. For example, configuration A1 ($D{=}1$, no refinement)
and configuration B1 ($D{=}2$, refinement enabled) achieve the same value gap
despite their different refinement settings. We observe analogous behaviour
for the $512\times512$ SW instance with $\PP_{\max}$, where the maximum
deviation across our configurations is $7.9\times 10^{-7}$.

  In contrast to the partitioning/refinement parameters discussed above,
  the boundary-change threshold $\eta_{\mathrm{thr}}$ is, one can argue, the most correctness-critical parameter.
  This stems from the fact that it essentially determines whether boundary values are propagated across the hierarchy or not.
  To investigate this, we sweep $\eta_{\mathrm{thr}}$ from $10^{-6}$ to $1.0$ across 17~values
  for each refinement depth $D\in\{1,2,5\}$, fixing $\theta{=}200$ and $8{\times}8$ partitions.
  Rather unexpectedly, the result is a sharp phase transition.
  All $\eta_{\mathrm{thr}}\le 0.9976$ yield the same gap ($1.3\times 10^{-4}$) regardless of depth,
  while all $\eta_{\mathrm{thr}}\ge 0.998$ produce a gap of $9.97\times 10^{5}$, reflecting the large cost assigned to states whose local
   sub-MDPs can no longer reach the goal through stale boundary values.
  The catch is of course that at $\eta_{\mathrm{thr}}{=}1.0$, boundary-triggered re-solving is effectively suppressed altogether and
  the algorithm terminates after only 2~global iterations.
  Even deep refinement ($D{=}5$) cannot compensate,
  because refinement alone does not substitute for boundary feedback.
 This observation is of course consistent with Theorem~\ref{thm:main}: when boundary feedback is suppressed, $\eta$ grows large and as a result the error guarantee degrades accordingly.

Hence, \acronym\ users should begin with $D=1$ or $D=2$ and $8 \!\times\! 8$ partitions, as these configurations perform well in diverse problems, while initial partitions larger than $12 \!\times\! 12$ typically degrade performance.
Deeper refinements (higher $D$) are justified mainly when observing large spreads in critical regions.
The boundary-change threshold $\eta_{\mathrm{thr}}$ should be set conservatively (e.g., $10^{-6}$) to ensure that boundary feedback is not suppressed (Figure~\ref{fig:eta-ablation}).

We should note that the preceding $8{\times}8$ experiments use partitions
well-matched to the grid structure, leaving little room for
refinement to improve quality.
To isolate the effect of refinement, we deliberately misalign the partition
by using a $3{\times}3$ grid on the 1024$\times$1024 MW instance ($\PP_{\max}$),
where vertical walls fall inside blocks, and cap local VI at 300~iterations
(Figure~\ref{fig:refinement-tradeoff}).
Without refinement ($D{=}1$, $D{=}2$), the policy gap stays at 21.6\%;
at $D{=}3$, it drops to 0.04\%.
Deeper refinement ($D{=}4$, $D{=}5$) yields diminishing returns, though,
improving the gap only to ${\approx}\,10^{-4}$\% and ${\approx}\,10^{-5}$\%
respectively.
Refinement is most valuable under a fixed compute budget with misaligned partitions; when the partition already
matches the problem structure, the gains are marginal.

\textbf{Guidance on using \acronym.}
  \acronym\ is most effective on models with limited cross-boundary dependencies (spatial or staged-progression MDPs). In practice, this means that transitions are mostly local within a block and the interaction across blocks is relatively sparse. However,
  on tightly coupled protocol models flat methods such as standard VI remain preferable.                                                 
  We emphasise that \acronym\ still produces correct results on such models; the limitation is purely one of efficiency and
  does not affect solution quality.
  A natural extension would be to detect when the hierarchical decomposition is slower than flat VI and fall back automatically.
  We also observed that initial partition size matters more than refinement depth or threshold: small grids ($8{\times}8$--$12{\times}12$) with
  $D\le 3$ yielded the best results.
  For new models, we suggest setting $\theta \in [0.2, 0.5]$ for probability objectives and $\theta$ scaled to the
  expected reward range for cost objectives (e.g., $200$--$500$ when expected costs are in the thousands).
  Finally, $\eta_{\mathrm{thr}}$ should be set conservatively (e.g., $10^{-6}$) so that significant boundary changes trigger re-solving
  of affected sub-MDPs (Figure~\ref{fig:eta-ablation}).

  \subsection{Threats to Validity}                                                                                                       
  \label{sec:threats}                                                                                                                    
  \noindent                                                                                                                              
  We mitigate \textbf{construct validity threats} that arise due to the lack of global context
  in hierarchically decomposed sub-MDPs by grounding \acronym\ in the well-established VI
  algorithm for probabilities and rewards. Furthermore, \acronym\ propagates values up the tree
  and supplies boundary values downward, while refinement is triggered whenever the value
  spread criterion is violated. This allows us to recover accuracy in those parts of the
  hierarchy where it is most needed.

  We limit \textbf{internal validity threats} by integrating \acronym\ into PRISM's explicit-state
  engine, reusing its core while supplying our algorithms for hierarchical partitioning,
  adaptive refinement, and local sub-MDP solving. By building on a widely tested code base, we
  minimise the risk that implementation bugs skew our measurements. Setting $\eta_{\mathrm{thr}}$ to a suitably small value, for
  example $10^{-6}$, bounds the boundary mismatch. However, the approach is not necessarily
  beneficial in all cases. For example, on tightly coupled MDPs the overhead may outweigh the
  gains (Table~\ref{tab:generalization}), while setting $\eta_{\mathrm{thr}}$ too permissively
  may suppress boundary feedback altogether (Figure~\ref{fig:eta-ablation}).

  We mitigate \textbf{external validity threats} by evaluating \acronym\ on large benchmarks
  that correspond to realistic warehouse robot scenarios of varying complexity and model
  sizes~\cite{fragapane2021planning}, and by comparing it against the VI and GS algorithms of
  PRISM~\cite{prism}. Assessing \acronym\ against other state-of-the-art tools such as
  Storm~\cite{storm} is not straightforward, for the reason that Storm is implemented in C++
  and employs different data structures and solver backends. Therefore, a fairer evaluation
  would require porting \acronym's core algorithm (Algorithm~\ref{alg:SHARP-generic}) to
  Storm's framework. However, \acronym's partition-refinement approach is solver-agnostic at the leaf level,
  meaning that any local solver could, in principle, replace VI. As a result, a
  Storm-based backend constitutes a natural direction for future work.

  Additionally, \acronym\ currently supports only single-objective properties ($\PP_{\max}$ and
  $\RR_{\min}$); multi-objective and constrained synthesis remain outside its scope.
  Lastly, more experiments on additional non-grid MDPs, and on additional application domains
  (e.g., self-adaptive systems~\cite{su2016iterative}, software product
  lines~\cite{profeat2016}) are needed to establish \acronym's applicability and scalability in
  domains beyond those in our evaluation.

\section{Related Work}
\label{sec:relatedwork}
\noindent
\textbf{Probabilistic models in Software Engineering.} There are numerous applications of probabilistic models in an SE context; for example~\cite{cheung2009phones} propose a framework using MDPs to dynamically optimise mobile phone software, while~\cite{ksentini2014cloud} apply a similar approach to optimise service migration decisions in the cloud for mobile users.
Similarly,~\cite{calinescu2011qos} uses probabilistic modelling for QoS management in service-based systems.
Moreover, \cite{evangelidis2018autoscaling, naskos2015scaling} use probabilistic model
checking to derive dependable auto-scaling policies for cloud deployments.
\acronym\ can, in principle, accelerate the policy synthesis step
underlying each of these applications when the state space
becomes prohibitively large.
\\
\textbf{Compositionality and abstraction--refinement for verification.}
CEGAR~\cite{clarke2000cegar} pioneered iterative refinement for non-probabilistic systems; probabilistic variants refine MDP abstractions with counterexamples~\cite{hermanns2008probabilistic,kattenbelt2010game}.
As discussed in Section~\ref{sec:into}, for hierarchical MDPs,~\cite{junges2022abstraction}
computes coarse probability bounds via parametric analysis and refines subroutines as needed.
Watanabe~\cite{watanabe2023compositional}
presents a compositional model checking framework for MDPs that uses the formal language of string diagrams to decompose models and compute optimal expected rewards.
Unlike these methods, \acronym\ partitions the concrete state space automatically and refines based upon value spread rather than counterexample elimination or formal compositional structure.
\\
\textbf{Search-based methods.}
Search-based approaches have also been used for related probabilistic synthesis problems. EvoChecker~\cite{gerasimou2015evochecker,gerasimou2018evo} uses evolutionary algorithms to synthesise probabilistic model configurations that satisfy QoS requirements, whereas EvoPoli~\cite{gerasimou2021evopoli} considers constrained multi-objective MDP policy synthesis and approximates Pareto-optimal policy sets. These methods target approximate search over configurations or policies. In contrast, \acronym\ assumes a fixed MDP and is concerned with accelerating exact single-objective synthesis by means of hierarchical decomposition and adaptive refinement.
\\
\textbf{Hierarchical RL.}
Early HRL frameworks such as Feudal RL~\cite{dayan1992feudal}, MAXQ~\cite{dietterich2000maxq},
and Options~\cite{sutton1999temporal} decompose an MDP into subtasks or macro-actions,
which significantly improves computational tractability.
Recent abstractions (e.g., AVI~\cite{jothimurugan2021abstractvi}) extend these ideas to continuous spaces.
Similarly, Monte Carlo-based approaches~\cite{bai2016hierarchical} treat abstracted MDPs as POMDPs
in order to scale tree search.
These methods target the reinforcement learning setting, where the transition model is unknown and must be explored, whereas \acronym\ operates in the model-based setting where the full MDP is given.
Note, however, that unlike \acronym, HRL methods usually learn from samples and they provide no optimality guarantees.

\section{Conclusion}
\label{sec:conclusion}

We presented \acronym, a hierarchical adaptive refinement approach for accelerating policy synthesis in large MDPs.
We implemented \acronym\ as an open-source extension to PRISM and evaluated it on scenarios with up to 1M states. The results show that \acronym\ can improve efficiency over PRISM-based policy synthesis using VI and GS, while preserving a high level of accuracy. We also provided formal guarantees showing that the composed policy remains near-optimal when the boundary mismatch is controlled. The approach is most beneficial for MDPs with spatial or staged-progression structure. On tightly coupled models, standard flat solvers may still be preferable from an efficiency point of view.

There are several directions in which this work can be developed further. For example, it would be useful to extend \acronym\ to multi-objective and constrained synthesis, and also to investigate how the approach behaves in multi-agent settings. In addition, more experimentation is needed in larger model instances in order to assess the scalability limits of the approach more thoroughly. Finally, it would be interesting to examine whether partition sizes and refinement thresholds can be selected automatically based upon the structure of the model, rather than being chosen manually.

\section*{Data Availability}
The SHARP implementation, benchmark models, experimental data, and supplementary proofs are available at \url{https://github.com/alexEvangelidis/sharp}.

\begin{acks}
This research was supported by the Horizon Europe projects AI4Work (101135990) and SOPRANO (101120990).
\end{acks}

\bibliographystyle{ACM-Reference-Format}
\bibliography{main-paper}

\end{document}